\documentclass[10pt,twocolumn,letterpaper]{article}

\usepackage{cvpr}
\usepackage{times}
\usepackage{epsfig}
\usepackage{graphicx}
\usepackage{amsmath}
\usepackage{amssymb}

\usepackage{enumitem}

\newcommand{\ms}[1]{{\color{blue}\bf [MS: #1]}}
\newcommand{\gb}[1]{{\color{magenta}\bf [GB: #1]}}
\newcommand{\todo}[1]{{\color{red} \bf [TODO: #1]}}

\usepackage[pagebackref=true,breaklinks=true,letterpaper=true,colorlinks,bookmarks=false]{hyperref}

\cvprfinalcopy 

\def\cvprPaperID{8441} 

\ifcvprfinal\fi
\begin{document}

\title{Improved Handling of Motion Blur in Online Object Detection}

\author{Mohamed Sayed \qquad \qquad \qquad Gabriel Brostow\\
University College London\\
\url{visual.cs.ucl.ac.uk/pubs/handlingMotionBlur/}
}

\maketitle
\thispagestyle{empty}

\begin{abstract}
We wish to detect specific categories of objects, for online vision systems that will run in the real world.
Object detection is already very challenging. It is even harder when the images are blurred, from the camera being in a car or a hand-held phone. Most existing efforts either focused on sharp images, with easy to label ground truth, or they have treated motion blur as one of many generic corruptions.

Instead, we focus especially on the details of egomotion induced blur. We explore five classes of remedies, where each targets different potential causes for the performance gap between sharp and blurred images. For example, first deblurring an image changes its human interpretability, but at present, only partly improves object detection. The other four classes of remedies address multi-scale texture, out-of-distribution testing, label generation, and conditioning by blur-type. Surprisingly, we discover that custom label generation aimed at resolving spatial ambiguity, ahead of all others, markedly improves object detection. Also, in contrast to findings from classification, we see a noteworthy boost by conditioning our model on bespoke categories of motion blur.

We validate and cross-breed the different remedies experimentally on blurred COCO images and real-world blur datasets, producing an easy and practical favorite model with superior detection rates. 

\end{abstract}

\begin{figure}[t]
\begin{center}
   \includegraphics[width=0.7\linewidth]{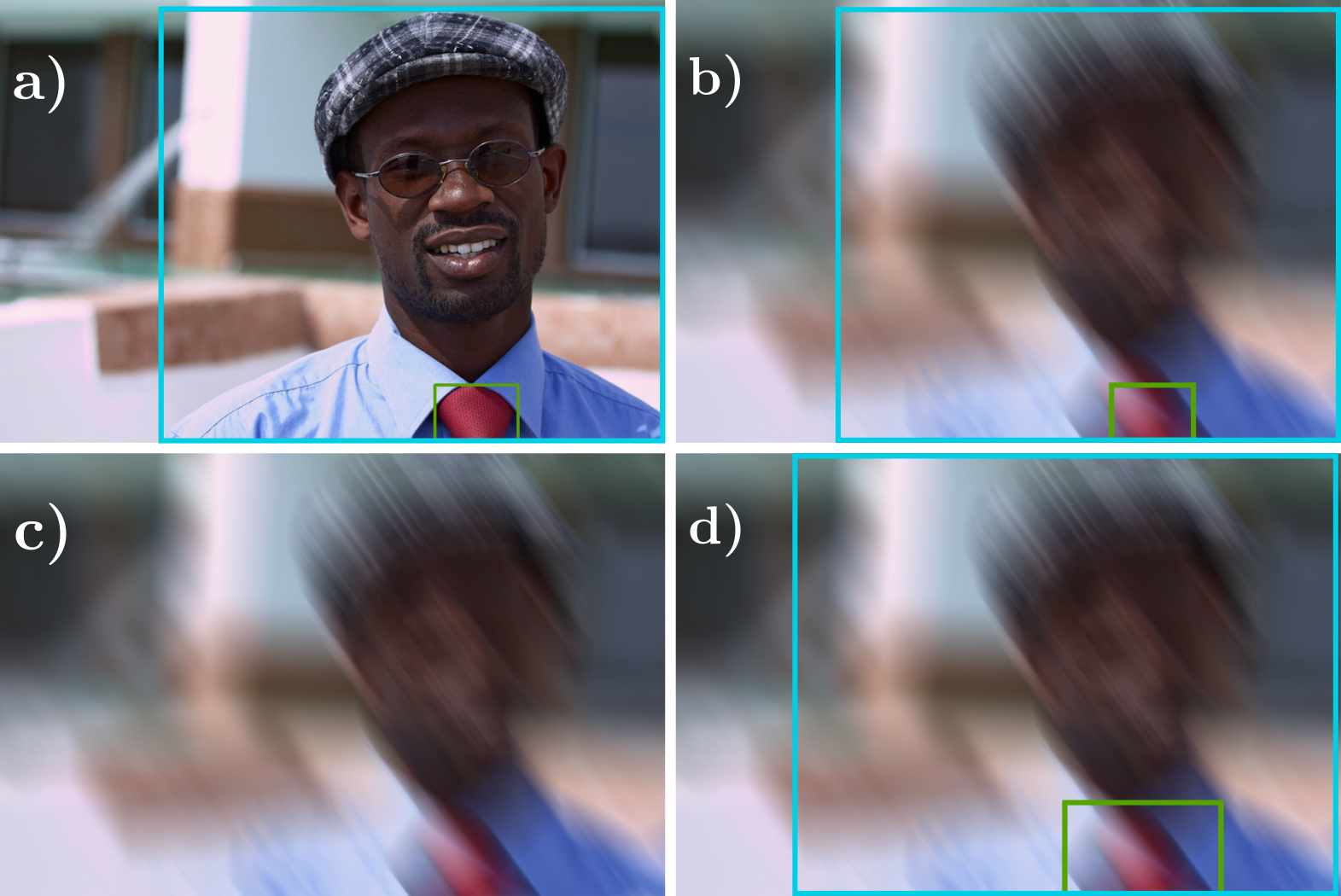}
\end{center}
\caption{a) Original sharp MS COCO~\cite{lin2014microsoft} image with object detections. b) Same image with significant linear motion-blur, with COCO ground-truth. c) Failed predictions from original Faster-RCNN. d) Predictions from network with our proposed model.}
\label{fig:teaser}
\end{figure}

\begin{figure*}[t]
\begin{center}
   \includegraphics[width=0.95\linewidth]{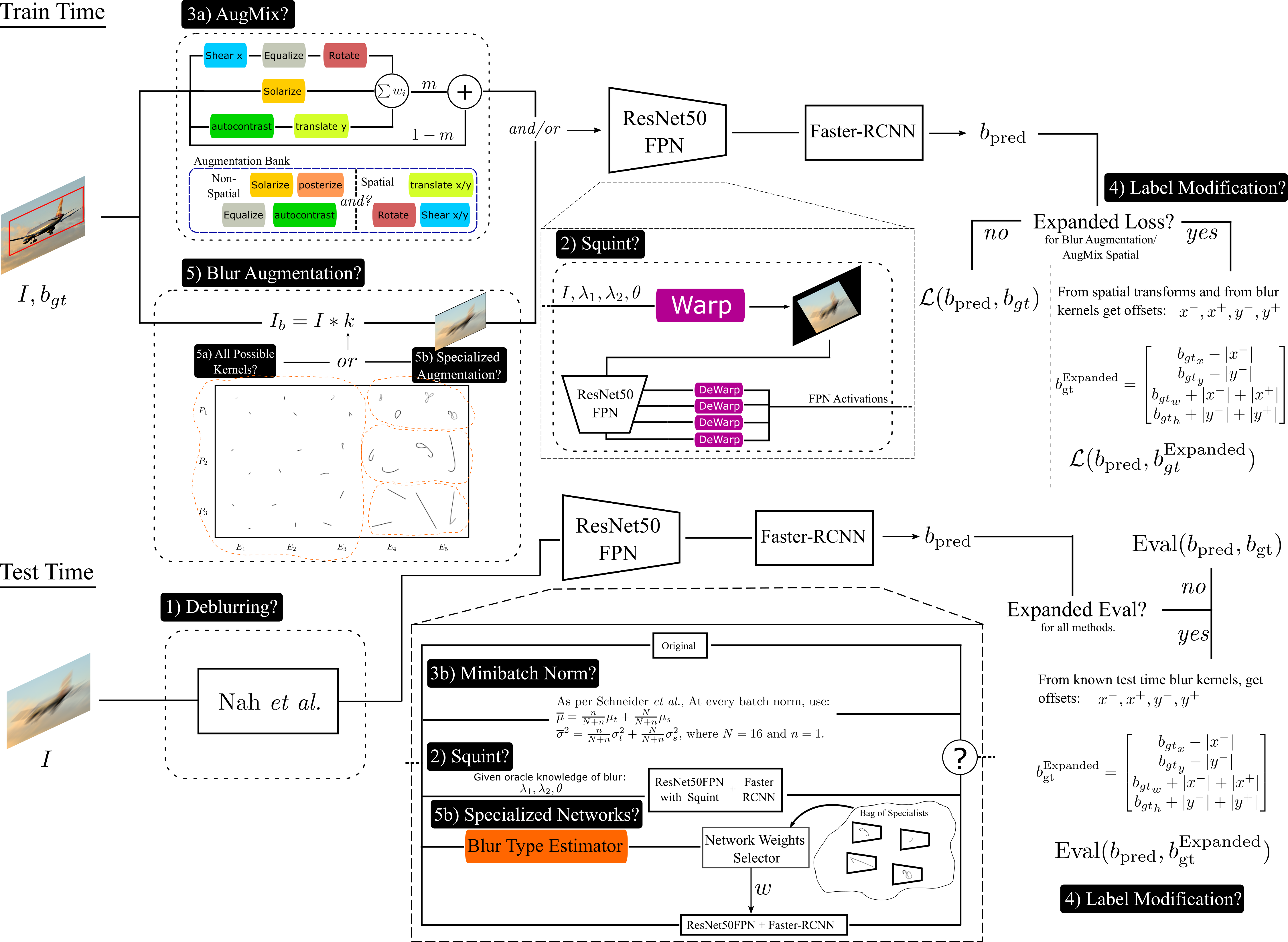}
\end{center}
   \caption{We explore the baseline with five categories of remedies across both train and test time. \textbf{1) Deblurring:} Deblur the input using Nah~\etal~\cite{nah2017deep} before passing it to the detector at test time. \textbf{2) Squint:} Influenced by~\cite{jaderberg2015spatial, recasens2018learning}, we undersample the incoming image on each axis, according to known blur kernel shape during training and testing (assuming oracle knows test kernel). We forward pass through the backbone and carry out the reverse sampling operation on outgoing activations, before passing them to the detection head. \textbf{3) Out-of-distribution:} Treat motion blur as an out-of-distribution robustness problem and use two leading methods as remedies: \textbf{3a) AugMix:} Augment training images as in AugMix~\cite{hendrycks2019augmix} with two main flavors: all augmentations from the AugMix paper, or only non-spatial augmentations. We then use the model normally at test time. \textbf{3b) Using minibatch statistics:} As in~\cite{schneider2020improvingCovariantShift}, modify the batch normalization statistics at every layer in the network to be a weighted average of the incoming minibatch (size one) statistics and the original training-averaged batch normalization statistics. \textbf{4) Label Modification} Under blur, the spatial extent and center of the bounding box are ambiguous. We experiment with training with labels expanded to include the superset of object locations given by the extent of the blur kernel's reach, under blur augmentation and when using spatial augmentation in AugMix. We also report results for both expanded COCO minival labels and without. \textbf{5) Blur Augmentation} Augment COCO images using either \textbf{5a)} a random selection of kernels across all possible motion-blurs we can generate, or \textbf{5b)} training networks specialized on specific blur-kernel varieties. At test time, we use either a network trained under general blur augmentation, or a system that incorporates a bag of models and a blur estimation module that selects the appropriate specialized network for the task.}
\label{fig:ModelOverview}
\end{figure*}

\vspace{-0.4cm}

\section{Introduction}
A little motion blur is present in most hand-held photography. Blur is ever harder to ignore because images are increasingly captured on the move, \eg by a gimbaled robot or from an autonomous vehicle. Precisely these on-the-go situations prompt us to explore: how much does motion blur severity impact object detection? What can be done about it? Detection is important because it underpins many other tasks, such as tracking and re-identification, and our initial scope is further narrowed to egomotion induced blur. 

Unsurprisingly, the severity of the blur correlates with detection failure~\cite{Bertasius_2018_ECCV}. Fig.~\ref{fig:teaser} shows an example. An ideal algorithm will make that degradation more gradual, and could someday enable a model that surpasses even a human's ability to see through blur. Instead of a single breakthrough, it is more likely that a combination of approaches is needed. Much like the ``devil in the details'' papers~\cite{Chatfeld11,Chatfeld14}, the task specifics and pipeline likely make a difference.

Our main contribution is an empirical exploration of five classes of remedies. These remedies are selected to cope with five proposed causes for reduced detection accuracy. The five cause/remedy pairs explored here are: \textbf{1)} \textit{Is the entire image too blurry to be useful?} Deblur test image first. \textbf{2)} \textit{Is texture mismatch along blur axes confusing the model?} Spatially transform image to compensate. \textbf{3)} \textit{Does test-time blur differ from training data?} Train model for out-of-distribution robustness, and/or perform test-time tuning of network. \textbf{4)} \textit{Are the training labels incorrect?} Customize labels to match detection-in-blur task and reconsider labels used for testing. \textbf{5)} \textit{Are egomotion blur types too diverse?} Treat detection in blur as a multi-task problem.

Overall, we propose a new model that focuses on the remedies from (4) and (5), and set a new standard for online object detection in egomotion-induced blur. 



\section{Related Work}

\textbf{Deblurring:} A close topic with valuable data and potential insights is image deblurring. The first canonical method for image deconvolution comes from Richardson~\cite{richardson1972bayesian} and Lucy~\cite{lucy1974iterative} where a known point spread function (PSF) - the blur kernel - is used to iteratively minimize an energy function to find a maximum likelihood estimate of the original image. Deblurring can be non-blind where the blur kernel is known~\cite{schmidt2013discriminative}, or it can be blind, where the kernel is either first estimated~\cite{singlePhotoDeblurring, learningToDeblur, kapuriya2019detection} - usually optimized with the final result~\cite{chen2019blind} - or the entire deblurring method is non-interpretable and runs end to end~\cite{nah2017deep, SRNDeblur}. Deblurring can also assume either a uniform blur kernel throughout an image~\cite{singlePhotoDeblurring, learningToDeblur} or variable nonuniform blur either due to camera egomotion (rotation, zoom)~\cite{whyte2012non}, depth-of-field effects~\cite{Tang_2019_CVPR}, or dynamic object motion blur~\cite{hyun2013dynamic, nah2017deep, kapuriya2019detection}.

Previous work has made use of an L0 sparse representation~\cite{L0rep}, dark image regions~\cite{darkPrior}, and multiple frames in a video~\cite{videoDeblur}. More invasive methods exploit hardware, including using a coded shutter~\cite{raskar2006coded}, inertial measurements~\cite{inertialDeblurringJoshi2010}, flash frame information integration~\cite{zhuo2010robustflash}, bursts of blurry images~\cite{aittala2018burst}, high and low frame rate cameras~\cite{tai2008imageHybrid}, or an event driven camera tied to an RGB sensor~\cite{pan2019EventDrivenBlur}.

Some deep learning deblurring methods are interpretable ~\cite{learningToDeblur, neuralDeblur, xu2017motion, nonuniformBlurRemoval}, but most are end-to-end~\cite{SRNDeblur, deblurGAN, dynamicSceneDeblurring, nah2017deep} with the recent state-of-the-art by Nah~\etal trained on the high frame rate GOPRO dataset~\cite{nah2017deep}. We explore using a state-of-the-art deblurring method as a preprocessing step, and measure the overall effectiveness of such a baseline.

Boracchi~\etal~\cite{boracchi2012modeling} generate a statistical model for motion blur kernel generation for benchmarking image restoration model performance. The blur kernels they generate are parameterized to simulate camera shake and exposure, making their kernel generation method a good candidate for synthesized blur augmented training.

Although deblurring's aesthetics driven approach means that there are competitive methods for extracting high frequency information in blur, use in an online vision application might be impractical, especially given that networks are very sensitive to changes in training distribution.

\textbf{Blur and Scene Understanding Tasks:}
Directly related to this work, Vasiljevic~\etal~\cite{vasiljevic2016examining} explore the effect of blur on ImageNet~\cite{imagenet_cvpr09} classification performance and blur augmentation strategies using a set of synthetically generated blur kernels; however they use a limited set of $100$ $17\times17$ pre-generated fixed length motion blur kernels and a restrictive image resolution of $384\times384$ during training and evaluation. They experiment with different blur types and fine grained blur augmentation for classification, but only a segmentation dividing blur types - and not across blur exposure with different kernel types - is considered; only defocus blur is explored for image segmentation. For image segmentation, they evaluate their networks using a soft boundary for accuracy, but do not explore the effect of spatial ambiguity on fine-tuning networks for blur during training, especially since only defocus blur (no shift in barycenter naturally) is used for fine-tuning the segmentation task. Vasiljevic~\etal note that knowing blur information apriori could be helpful, but don't explore such a blur estimator.

Overall, we find that building explicit robustness into vision models for dealing with realistic camera motion blur needs more exploration, especially for spatial tasks.

\textbf{Out-of-Distribution Robustness:} Recent work~\cite{hendrycks2019benchmarking, hendrycks2020many} treats image corruptions (brightness, contrast, snow, noise, blur) as out-of-distribution samples compared to the in-distribution clean images a network was trained on. ImageNet-C~\cite{hendrycks2019benchmarking} is a variant of the ImageNet classification dataset that contains images corrupted by 15 different types of canonical image corruptions, and is used as a benchmark for out-of-distribution model performance. Crucially, ImageNet-C - and others such as ImageNet-R~\cite{hendrycks2020many} and ImageNet-A~\cite{hendrycks2019natural} - are not meant to be trained on. Instead, the argument is that a model's ability to generalize to images outside of the training set's distribution can be measured by evaluating its performance on these datasets. Although ImageNet-C contains motion blur corruptions, the method only considers straight line  motion blur kernels. Michaelis~\etal~\cite{michaelis2019benchmarking} use the same corruptions from ImageNet-C to produce a robustness benchmark for detection, by augmenting MS COCO~\cite{lin2014microsoft}. COCO-C also includes straight line blurred images, but changes in labels under the spatial ambiguity brought upon by blur are not addressed. We call this type of naive blur 'Non-Centered,' and we show why it's important for spatial reasoning.

In the data augmentation space, AutoAugment~\cite{cubuk2019autoaugment} finds an optimal augmentation policy for a model and dataset pair achieving state-of-the-art accuracy for classification datasets, but requires 15,000 compute hours on an NVIDIA Tesla P100 for training ImageNet. Rusak~\etal~\cite{rusak2020NoiseAdver} propose an adverserial noise training scheme for increasing classification model accuracy and robustness on ImageNet-C, mainly combating pixel noise and not blur. AugMix~\cite{hendrycks2019augmix} is an augmentation strategy for improving classification model robustness to out-of-distribution images. It involves alpha blending copies of a training image that have been corrupted by a random chain of image augmentations. They use the same corruptions in~\cite{cubuk2019autoaugment}, including both pixel level value changes and spatial augmentations. Although the AugMix paper also doesn't discuss how spatial augmentation should affect spatial labels, we explore the effectiveness of AugMix for blur robustness after making decisions on how spatial labels should be changed.

Schneider~\etal~\cite{schneider2020improvingCovariantShift} analyze the effect of normalizing activations in batch normalization layers using a weighted average of the statistics of both the source training set and the minibatch. Their method achieves state-of-the-art on ImageNet-C and improves ImageNet-C robustness on vanilla Resnet-50 classification models, even with a minibatch of size one.

While these methods are a promising way of increasing model robustness to unseen corruptions, the aim of the proposed work is to explore the specific impact of motion blur on detection, and so we focus our effort on manufacturing the most realistic blur kernels available in the literature.

\section{Designing Detection Models for Motion Blur} \label{Sec:DesigningModelsSetup}
To improve online object detection, we propose a unified framework that allows us to measure the impact of different remedies and their combinations. The framework is based on a state-of-the-art object detector, Faster-RCNN~\cite{ren2015faster}, with training and testing on data derived from the MS COCO~\cite{lin2014microsoft} detection dataset. The baseline and data are explained in this section. 
The proposed remedies are explained in detail in Sec.~\ref{Sec:Remedies}, and are evaluated in Sec.~\ref{Sec:Experiments}. 
Figure~\ref{fig:ModelOverview} illustrates both the baseline model, and the different enhanced alternatives.

\subsection{Detection Baseline}
For reproducibility, we use the pretrained Faster R-CNN variant trained on COCO, available through Pytorch's torchvision~\cite{pytorch} library, as a baseline for all our experiments. We use a ResNet-50~\cite{resnet} backbone with a Feature Pyramid Network (FPN)~\cite{fpnBackbone}. This baseline achieves 58.5 mAP@0.5 and 37.0 mAP@0.5:0.95 on the COCO test set. While other models achieve better accuracy on the COCO minival set, we choose this framework for its accessibility and as a good baseline representation of a canonical detection framework with top-10 performance for the backbone's size~\cite{jiao2019survey, liu2020deep}.

\subsection{Selecting Data for Training and Testing}

Ideally, we'd select data with detection labels for images exhibiting motion blur. Due to the way MS COCO is gathered~\cite{lin2014microsoft}, there are very few blurry images in the dataset. This leaves us with the task of generating synthetically blurred COCO images for both training and evaluation. Related but not directly applicable here, there are multiple real-world image datasets for deblurring. These were generated using high frame rate video~\cite{nah2017deep} or shutter tied cameras~\cite{rim_2020_ECCV}. They either don't contain enough images for training and evaluating detection models (\cite{rim_2020_ECCV} only contains 5500 images) and/or lack object annotations. Zhang~\etal~\cite{zhang2020deblurring} generate blurry images as part of a GAN architecture for deblurring. Although they train the blur generation module using a discriminator trained on real world blurry images, it is not trivial to modify labels, given spatial ambiguity, since camera motion is not made explicit. Brooks \& Barron~\cite{brooks2019learning} use multiple adjacent images (as few as two) to generate realistic motion blur. But to use that would require a video or stereo dataset with ground truth labels for the detection task. 

This leaves methods that synthesize blurry images via convolution with synthetic motion blur kernels~\cite{schmidt2013discriminative, boracchi2012modeling, vasiljevic2016examining, michaelis2019benchmarking, hendrycks2019benchmarking}. ImageNet-C~\cite{hendrycks2019benchmarking} and COCO-C~\cite{michaelis2019benchmarking} contain images blurred using straight line motion blur exclusively, with no control over simulated camera shake. Vasiljevic~\etal~\cite{vasiljevic2016examining} use a limited set of motion-blur kernels since they are constrained by a fixed length spline formation model. Boracchi \& Foi~\cite{boracchi2012modeling} describe a method that allows control over different characteristics of a camera's trajectory through space, including the amount of shake and jerk with variable exposure.


\subsection{Blur Generation and Space Discretization}

\begin{figure}[t]
\begin{center}
   \includegraphics[width=0.8\linewidth]{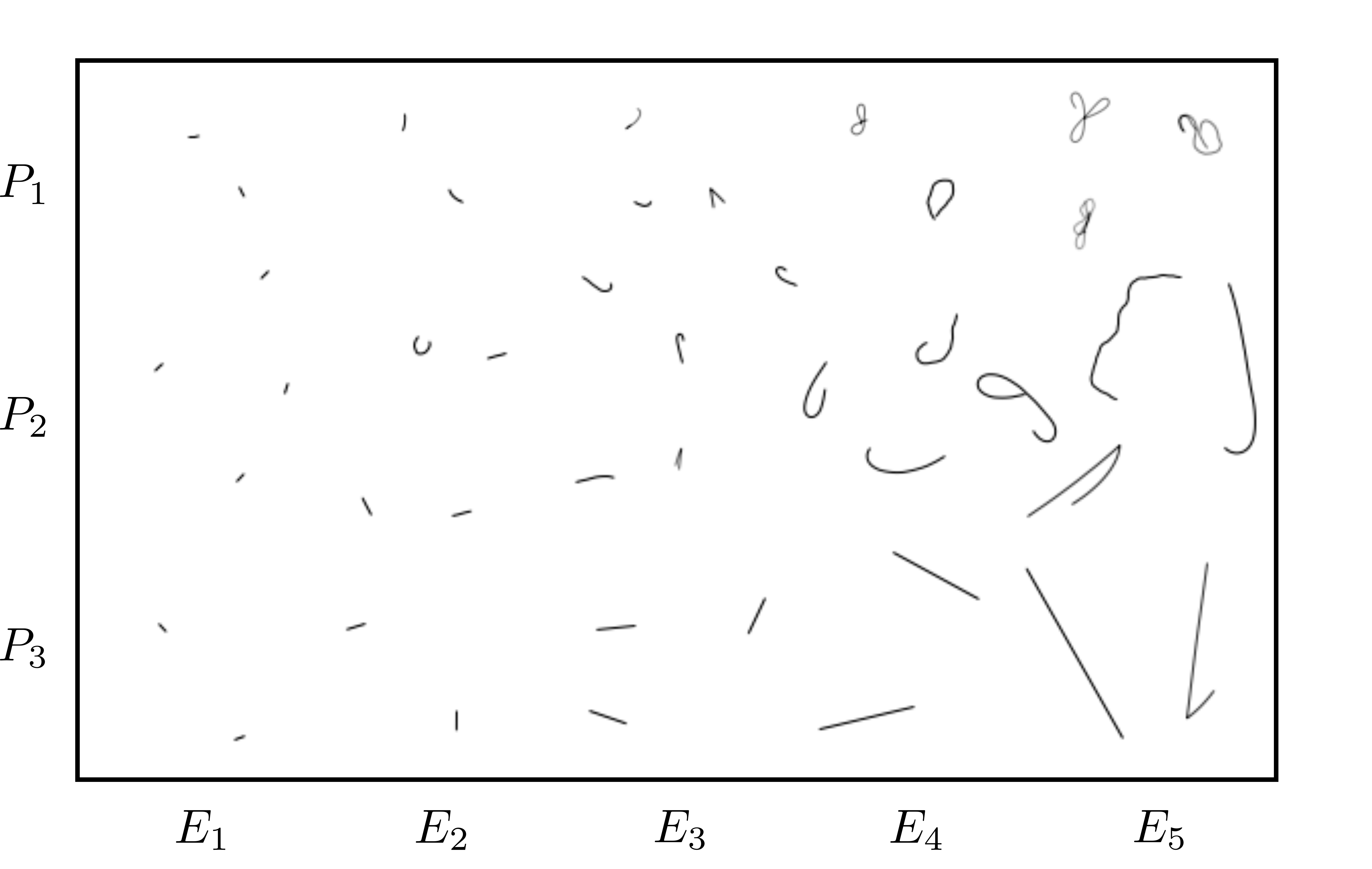}
\end{center}
\caption{Example blur kernels based on Boracchi \& Foi's~\cite{boracchi2012modeling} motion-blur model. All kernels occupy a space of $128\times128$. In our modification, used everywhere unless stated otherwise, kernels are centered by shifting the barycenter of all nonzero points to the origin of the filter. $P_1$ type kernels tend to be very erratic, while $P_3$ kernels contain mostly rectilinear trajectories representative of when camera ego motion is linear. 
}
\label{fig:examplePSFs}
\end{figure}

We adapt the blur kernel generation method from Boracchi \& Foi~\cite{boracchi2012modeling}. We fix their high level controllable parameter $P$ to one of three values, $P_{1-3}$, representing three distinct types of camera motion. We also modulate exposure via early camera trajectory clipping.  



First, we generate a trajectory by finding a random path in 2D space. We assign an initial velocity vector $\mathbf{v_0}$, drawn at random from a unit circle, and a position in space $\mathbf{x_0}$ for the camera. At every step, the camera's velocity vector is updated by the acceleration vector, 

 \vspace{-10pt}

\begin{equation}
\begin{aligned}
\mathbf{\Delta v} &= P (\mathbf{\Delta v_g} - I\mathbf{x_t}),
\end{aligned}
\end{equation}

\noindent where $\mathbf{\Delta v_g}$ is random acceleration with elements drawn from $\mathcal{N}(0,\,\sigma^{2})$. $I\mathbf{x_t}$ is an inertial tendency for the camera to stay where it is, and $P \in P_{1-3}$ is the high level anxiety parameter we fixed above. $P_3$ has the highest random velocity change on every step. Further, to model a camera jerk, with a randomly sampled indicator function, 
the acceleration update also includes a component equal to twice the current velocity vector in a random direction, so

\vspace{-10pt}

\begin{equation}
\begin{aligned}
\mathbf{\Delta v} &= \mathbf{\Delta v} + 2P\lvert\mathbf{v}\rvert\mathbf{\Delta v_j},
\end{aligned}
\end{equation}

\noindent where $\mathbf{\Delta v_j}$ is sampled from the unit circle. Again, with a high P, there is a higher chance of a jerk happening and a shakier camera. When starting a trajectory, $I$, $\sigma^2$, and $j$ are drawn once from uniform random distributions to increase variability under the same class of blur $P \in P_{1-3}$. Note that this leads to some overlap between kernels generated across different $P$s.

In summary, the type of camera behavior falls into one of three classes: 1) $P_1$ simulates a very nervous camera, 2) $P_2$ for back and forth behavior, and 3) $P_3$ simulates mostly straight rectilinear motion-blur. To simulate exposure, we stop the motion path early using the exposure factor (trajectory length) $E$. We discretize exposure to one of $5$ values, $E_{1-5}$. Examples of these kernels can be seen in Fig.~\ref{fig:examplePSFs}. Subpixel interpolation produces kernels for convolving sharp images. 

\subsection{Implementation Details}

\textbf{Kernel Generation:} To speed up training, a corpus of 12,000 blur kernels is generated for every pair of ${\{P_{1-3}\} \times \{E_{1-5}\}}$, for a total of 180,000 possible motion blur kernels. However, random kernels are generated on the fly during evaluation for each combination of blur type and exposure, with fixed seeds for reproducibility. Trajectory length is 96 and blur kernels fit in $128 \times 128$ filters. 

\textbf{Blurring:} Unlike in ~\cite{hendrycks2019benchmarking, vasiljevic2016examining, michaelis2019benchmarking}, we don't resize images to a fixed size before blurring. Instead each image is convolved separately with reflection padding, to account for what would otherwise be real world data. We opt not to resize our blur kernels to match image size as a way of simulating changes in focal length. We implement sparse convolution on the GPU for applying blur kernels. As per Sec.~\ref{subsec:customlabels}, we make sure to center our motion-blur kernels by translating their barycenters to the center of the filters.

\textbf{Training:} All networks start from a base Resnet-50FPN pretrained on COCO. We use an FPN framework that outputs activations at four scales from the backbone. There was no apparent difference in blur augmented performance when training all five blocks \vs fixing the weights of the first two. 

\begin{figure*}[t]
\begin{center}
   \includegraphics[width=0.9\linewidth]{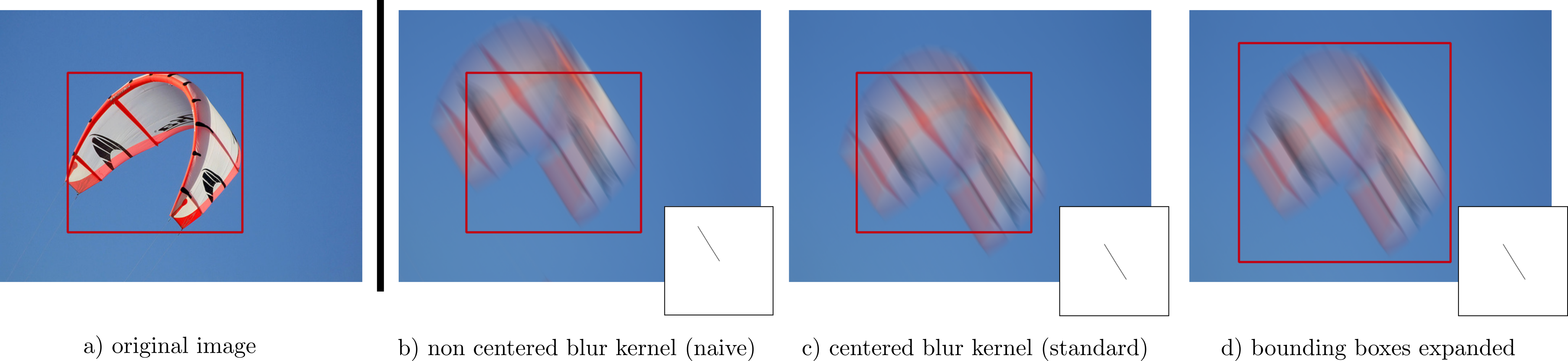}
\end{center}
   \caption{a) An image from the MS COCO~\cite{lin2014microsoft} trainset with an associated bounding box label. b) The same image but blurred and with the same non-translated bounding box, now introducing a training/evaluation mismatch. The object may have been there at the start of the exposure (or end if the kernel smeared the other way), but this is certainly not true for the rest of the image. c) The first remedy, centering the kernel by shifting the barycenter of the nonzero points to the center of the filter. d) The object is smeared outside of the original COCO bounding box, so the label is expanded using the filter's max/min points to capture the superset of object locations. }
\label{fig:kernelAndBoundingBoxHandling}
\end{figure*}

\section{Proposed Remedies for Improved Detection} \label{Sec:Remedies}
Suspecting specific underlying causes for the adverse effects of blur on detection, we now propose bespoke remedies. 
Where appropriate, some of the remedies are also crossbred, and experimental results appear in Sec.~\ref{Sec:Experiments}.

\subsection{Deblurring as a Pre-process} 
Image deblurring is useful for aesthetic purposes, but could also aid other vision tasks. To test this remedy, we use the recent deblurring model from the GoPro dataset paper,~\cite{nah2017deep}, before passing the result to the detector. Deblurring is a slow process, by $12\times$ in this case, so heavy optimizations would be needed for an online robot. 

\subsection{Reconciling Texture Information With Scale} \label{Squint}
When motion-blur is biased to one major direction over another, it removes more high frequency information (and texture) in that direction. It is reasonable to expect a network is not natively designed for this imbalance. CNNs usually understand texture and shape information across multiple scales under the same aspect ratio, but we're also asking the network to deal with a texture imbalance along the blur kernel's major axis. Influenced by the work on Spatial Transformers~\cite{jaderberg2015spatial} (which proved a slightly inferior baseline) and neural sampling layers~\cite{recasens2018learning}, we instead undersample the incoming image along the principal components of the blur kernel. The reverse  operation is carried out, using reciprocal scaling factors, on every activation output from the backbone. This ``Squint'' process is done at both train and test time.  For best-case testing, an oracle is assumed to know the blur kernel.

\subsection{Training \vs Test Distribution} \label{trainvstest}
We consider treating complex motion-blur as an out-of-distribution corruption as in~\cite{hendrycks2019benchmarking, michaelis2019benchmarking, hendrycks2019augmix, schneider2020improvingCovariantShift}, and use two promising methods from the OOD literature. We use AugMix~\cite{hendrycks2019augmix} as a training time remedy. We propose three flavors, the first is a purely pixel level version where we augment pixel intensities only. The second applies all spatial augmentations as well as suggested in~\cite{hendrycks2019augmix}, but does not reconcile the shifts in bounding box changes. The third is an ``Expanded'' version following Sec.~\ref{subsec:customlabels}, where we change COCO labels at train time to match the superset of where an object is shifted to across branches. AugMix roughly approximates blur when augmentations are selected that translate an image before concatenating with other branches.

Further, at test time, we use covariate shift adaptation from the upcoming~\cite{schneider2020improvingCovariantShift}. The first step is to get a weighted average of the incoming activation statistics of the minibatch ($n=1$ for online inference) and the source statistics of the model where $N=16$, 


\begin{equation}
\begin{aligned}
\overline{\mu} = \frac{n}{N + n}\mu_t + \frac{N}{N + n}\mu_s, \text{and}
\end{aligned}
\end{equation}

\vspace{-0.25cm}

\begin{equation}
\begin{aligned}
\overline{\sigma}^2 = \frac{n}{N + n}\sigma^2_t + \frac{N}{N + n}\sigma^2_s.
\end{aligned}
\end{equation}

We then use these new normalization statistics for batch normalization in all network layers.

\subsection{Customizing Labels} \label{subsec:customlabels}
When an image is motion blurred, 
objects are no longer confined to the bounding boxes they had occupied in the sharp image. The objective may no longer be to estimate that original bounding box. 
See Fig.~\ref{fig:kernelAndBoundingBoxHandling}. We discuss two remedies for this problem, that apply when training under augmentation and for evaluation.

\textbf{Kernel Centering} \label{kernelCentering}
The start point of a motion blur path corresponds to the exposure at $t=0$. 
Any path that leads away from the center, as in Fig.~\ref{fig:kernelAndBoundingBoxHandling}(b), will offset the blurred version of the object in some direction, so the ``ground truth'' bounding box is no longer centered on the blurred object. 
This introduces a mismatch between the blurred input and its label. 
This mismatch is created in ~\cite{hendrycks2019benchmarking, michaelis2019benchmarking}, introducing label ambiguity and noise in training. 

We 
center a kernel using a weighted average of the kernel's nonzero points. The aim is to have the detection framework learn to localize objects based on where they are, on average, during the exposure. This remedy is similar to how \cite{rim_2020_ECCV} aligned images from paired long/short exposure cameras to train for deblurring. In our case, the training loss is noisier when training on non-centered kernels, and the drop in accuracy can be up to 8-10mAP@50 (see "Non-Centered Augmented" and "Standard Augmentated" in Fig.~\ref{fig:nonExpandAccuracyValues}) points with the most severe blur. All networks shown here will be trained and evaluated with centered kernels, except when explicitly mentioned. We include ablation experiments with no-centering in the supplemental.

\textbf{Expanding Target Boxes}
Compared to the original bounding box, the expanded label can cover the superset of pixels where an object projected during an exposure; see Fig.~\ref{fig:kernelAndBoundingBoxHandling}(d). 
A worst-case scenario could occur without this correction during training: for a small object, the sharp image's label could seemingly miss the blurred object entirely due to IOU cutoffs. As a remedy, for every generated centered kernel, we find the maximum offsets for non zero kernel elements in both 2D axes, ${x^{-}, x^{+}, y^{-}, y^{+}}$, and use them to expand the boundaries of COCO bounding box labels. The new bounding box labels (top left and width/height) are now

\begin{equation}
\begin{aligned}
\hat{b}_x &= b_x - \lvert x^{-} \rvert \\
\hat{b}_y &= b_y - \lvert y^{-} \rvert \\
\hat{b}_w &= b_w + \lvert x^{-} \rvert + \lvert x^{+} \rvert \\
\hat{b}_h &= b_h + \lvert y^{-} \rvert + \lvert y^{+} \rvert.
\end{aligned}
\end{equation}

We train variants of our networks with these expanded boxes alongside kernel centering. During test time, we evaluate these networks using expanded bounding boxes.

\subsection{Specializing for Categories of Blur} \label{blurSpec}
The final category of remedies explores if egomotion induced blur is perhaps multiple problems masquerading as one.
We explore training blur specialized networks on specific partitioned segments of the blur space, as if categories of blur are mutltiple distinct tasks. The findings on recognition in~\cite{vasiljevic2016examining} show that specialized networks can \emph{sometimes} achieve higher task accuracy on their respective blur types than general blur augmented networks. 

\textbf{Two Specialized Meta-Models} We make two sets of specialized networks, that differ in how the motion kernels are clustered into categories. 
First, motion blur is grouped based on the type of kernel $P$, alone, leading to a bespoke network for each of $P_{1-3}$ with a fourth generalist network trained for all types and exposures. 

%
%
%
%
%
%
%

The second grouping creates three networks specialized at each $P$ but exclusively on long exposure blur. One further network handles all low exposure blur. As per~\cite{geirhos2018imagenet}, networks are biased toward texture. Instead of using this knowledge to create more corruption robust networks, it is exploited here to make more shape biased networks for substantial motion blur.

\textbf{Blur Estimation and Network Selection} A ResNet-18 blur estimator module is added, and runs $10\times$ faster than the detection framework. The estimator categorizes the blur present in the image at test time. 
One network is trained on 16 classes (sharp and the combinations of all exposures and blur types) and the other network focuses on the separation between specific blur types at high exposures and general blur at low exposure (four classes). Details of how these estimators are trained are in the supplemental materiel. 

\section{Comparisons and Evaluation} \label{Sec:Experiments}

\begin{figure*}[t]
\begin{center}
\includegraphics[width=1.0\linewidth]{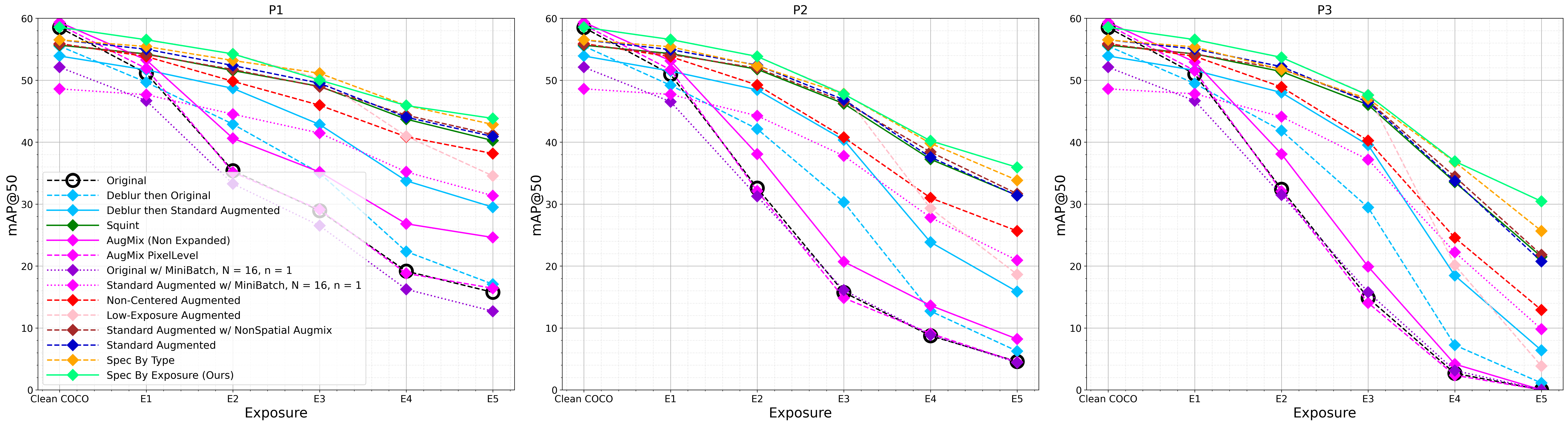}
\end{center}
   \caption{Non-expanded Standard Labeling COCO minival mAP@50 accuracy values across the three blur types and then exposures. Almost all methods and hybrids improve beyond the Original network. There is no benefit in augmenting for blur, and then using either minibatch statistics or a deblurer network.}
\label{fig:nonExpandAccuracyValues}
\end{figure*}

\begin{figure*}[t]
\begin{center}
\includegraphics[width=1.0\linewidth]{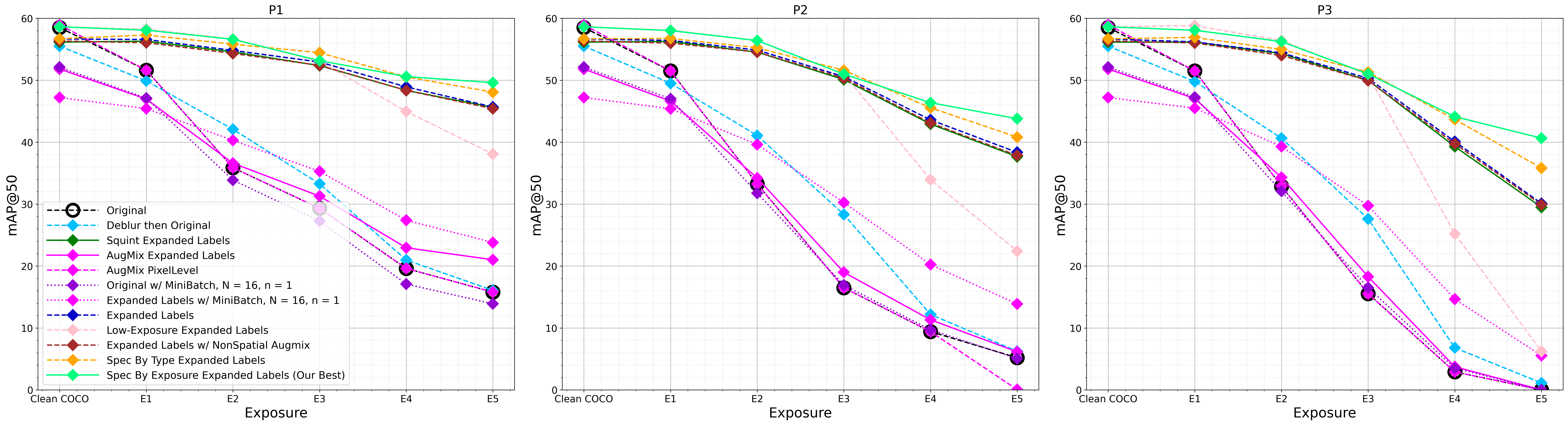}
\end{center}
   \caption{Expanded Labeling COCO minival mAP@50 accuracy values across the three blur types and then exposures. The Spec by Exposure network (``Ours'') excels at both ends of the exposure extremes, likely due to the biased training and specialized networks it enjoys. All expanded-box trained networks (except AugMix trained on expanded labels via spatial augmentation) perform better than their standard counterparts. Networks augmented with both blur and non-spatial AugMix perform well at low exposures and dataset generalization; in Table~\ref{table:realBlurData} we show performance for Spec By Exposure that makes use of this for improved performance on other datasets.\iffalse Note that the network augmented with both non-spatial AugMix and blur augmentation (Brown) maintains accuracy for sharp images too, where other blur augmented networks slump.\fi}
\label{fig:expandAccuracyValues}
\end{figure*}

\begin{table*}[h]
\begin{center}
\begin{tabular}{|l||c|c||c|c||c|c||c|c|}
\hline
& \multicolumn{2}{|c||}{GOPRO~\cite{nah2017deep}} & \multicolumn{2}{|c||}{RealBlur~\cite{rim_2020_ECCV}} & \multicolumn{2}{|c||}{REDS~\cite{Nah_2019_CVPR_Workshops_REDS}} & \multicolumn{2}{|c|}{GOPRO Expanded*} \\
\hline
Model & Sharp & Blurry & Sharp & Blurry & Sharp & Blurry & Sharp & Blurry\\
\hline
Spec by Exposure w/ \textdagger&34.89&\textbf{28.15}&43.23&\underline{36.37}&39.86&\textbf{32.55}&      38.06&30.99\\
Low-Exp Net w/ NonSpatial AugMix \textdagger&34.89&\underline{27.87}&43.23&36.28&39.86&\underline{32.53}&38.06&30.92\\
Spec by Exposure&34.80&27.05&42.91&35.26&40.06&31.51&                                       36.79&\underline{31.14}\\
Spec by Type&32.42&27.50&40.91&36.11&36.96&30.33&                                                       35.26&31.01\\
Low-Exposure Net&34.80&26.91&42.91&35.20&40.06&31.47&                                          36.79&\textbf{31.60}\\
Standard Aug w/ NonSpatial AugMix\textdaggerdbl&32.16&27.05&42.77&\textbf{36.82}&35.61&28.72&           35.18&29.94\\
Standard Augmented &32.42&26.54&40.93&35.63&36.96&29.24&                                                35.26&30.75\\
Non-Centered Augmented&33.08&25.66&40.47&34.47&36.82&28.80&                                             36.74&28.47\\
Standard Aug w/ Minibatch &30.06&23.78&35.01&31.32&28.99&24.94&                                         33.32&26.79\\
Original w/ NonSpatial AugMix &33.68&19.10&43.90&29.63&43.38&23.52&                                     36.63&22.16\\
Original w/ Minibatch &31.78&16.73&38.54&27.17&36.03&22.18&                                             33.45&20.50\\
Deblur then Standard Augmented&12.53&5.14&32.53&28.90&34.53&28.42&                                      34.54&29.72\\
Deblur then Original&10.58&2.24&31.45&28.76&40.56&31.84&                                                35.74&26.70\\
\hline
Original &35.85&19.64&42.39&29.00&42.02&24.38&                                                          36.33&22.30\\
\hline

\end{tabular}
\end{center}
\caption{mAP@0.5 for models trained on COCO images with synthesized blur and evaluated on real world blur datasets using predictions on sharp images using DetectoRS~\cite{detectors} as groundtruth. While NonSpatial AugMix doesn't improve performance for blur augmentation on the COCO minival, it does increase the performance of the COCO trained Low-Exposure\textdagger~and Standard Augmented\textdaggerdbl~networks on real world datasets. Spec by Exposure w/\textdagger~ utilizes the NonSpatial AugMix augmentation version of the Low-Exposure network. *Expanded augmented trained versions of the networks are used on this expanded labels test-set.}
\label{table:realBlurData}

\end{table*}

We report COCO minival results for all proposed remedies, at both test time and train time. Detection accuracy at mAP@50 is reported for all below-listed models and variants in Fig.~\ref{fig:nonExpandAccuracyValues}  and Fig.~\ref{fig:expandAccuracyValues}, where the former uses COCO original labels, and the latter uses expanded labels. We also report accuracy results on two pseudo-real blur datasets, GOPRO~\cite{nah2017deep} and REDS~\cite{Nah_2019_CVPR_Workshops_REDS}, and a real-world blur dataset, RealBlur~\cite{rim_2020_ECCV}, obtained using shutter tied cameras. These datasets don't have box annotations, so we utilize a state-of-the-art high accuracy detector, DetectoRS~\cite{detectors}, to obtain pseudo-groundtruth bounding-boxes for evaluation. For evaluating expanded bounding boxes, we generate our own GOPRO testset using grountruth sharp frames and use flow computed using ~\cite{teed2020raft} for bounding-box expansion.

Names in the figures are explained below, and map to the five remedy categories. Qualitative video here: \url{visual.cs.ucl.ac.uk/pubs/handlingMotionBlur}. 
\begin{itemize}[noitemsep]
\item \textbf{Standard Augmented} and \textbf{Expanded Labels} were trained on non-expanded but centered COCO labels, and expanded and centered COCO labels respectively. Both were trained on a 10/90 mixture of sharp to blurry images across all blur types. 

\vspace{0.1cm}

\item \textbf{Deblur then original} and \textbf{Deblur then Standard Augmented} are both modes of operation where the image is first deblurred using~\cite{nah2017deep} then run through either the original network or a \textbf{Standard Augmented} network respectively.

\vspace{0.1cm}

\item \textbf{Squint} and \textbf{Squint Expanded Labels} come from Sec.~\ref{Squint}, and have been trained either using standard labels under blur or expanded labels, respectively.

\vspace{0.1cm}

\item \textbf{AugMix~\cite{hendrycks2019augmix}} Hendrycks~\etal~\cite{hendrycks2019augmix} As described in Sec.~\ref{trainvstest}, we evaluate a non spatial version, \textbf{AugMix PixelLevel}, a spatial version without label expansion, \textbf{AugMix}~\cite{hendrycks2019augmix}, and a version trained with expanded labels as per augmentations and evaluated with blur based label expansions \textbf{AugMix Expanded Labels}.

\vspace{0.1cm}

\item \textbf{Standard Augmented w/MiniBatch} and \textbf{Expanded Labels w/MiniBatch} follow Schneider~\etal~\cite{schneider2020improvingCovariantShift} and use modified minibatch normalization with $N=16$ and $n=1$ as in Sec.~\ref{trainvstest} with networks that have been augmented for blur using either standard labels or expanded labels respectively.

\vspace{0.1cm}

\item \textbf{Standard Augmented w/ NonSpatial AugMix} and \textbf{Expanded Labels w/ NonSpatial AugMix} have been trained by first transforming the image using non spatial AugMix then blurring the image and training with expanded labels. NonSpatial AugMix augmentation helps when generalizing to other datasets at low exposure blur. 

\vspace{0.1cm}

\item \textbf{Spec by Type} is the first bag of specialists from Sec.~\ref{blurSpec} where the \textbf{Standard Augment} network is used when no blur is detected. \textbf{Spec by Exposure} is the second bag of specialists where each of three networks specializes in one $P$, but only at long exposures, and one network handles all short exposures and sharp images, \textbf{Low-Exposure Augment}. Again a blur estimator trained for these classes selects the right network. \textbf{Spec by Type Expanded Labels} and \textbf{Spec by Exposure Expanded Labels} are obviously variants trained on expanded labels. These networks and associated mode of operation outperform the rest due to their exploiting of the texture \vs bias trade-off and the use of an accurate blur estimator. Notably, the network responsible for sharp and low exposure blur recovers the accuracy lost on sharp images usually associated with blur augmentation networks.

\end{itemize}

\section{Discussion}
  
\vspace{-0.2cm}

We achieve state-of-the-art object detection results for egomotion-blurred images. We have succeeded in identifying that two factors adversely affect detection in such images. The first is that labels for sharp images should be customized for the motion-blur domain. In our remedies, that means translating and expanding the bounding box labels to match the blurred versions of relevant objects. The second is that categories of motion blur are distinct enough for the model to be trained for each blur-category separately. Interestingly, the second factor is the opposite of what \cite{vasiljevic2016examining} found with recognition tasks, where mixing blur-types during training was effective.

Through our ``differential diagnosis'' approach, the other three factors explored here seem unpromising for explaining the destructiveness of blur on CNN-based detection. These negative results are not conclusive, as the remedies may simply be immature. For example, better deblurring may eventually restore missing texture at all scales.

In the future, to reduce the memory footprint of our favored solution, the blur-selector and distinct exposure-specific models could be combined into one multi-task model. They are already end-to-end differentiable, but then they could share layers. 
Further progress in this direction could benefit from a distilled dataset that allows for detection labels and blur from real data, perhaps through the use of event driven cameras or multi-camera datasets. 

One clear limitation is that even sharp images have~$<~60~mAP@0.5$ detection accuracy with a realtime-capable backbone, before blur hurts the situation further. Depth or disparity data would help address scenes with dynamic blur, since the blur kernels are depth-dependent. 360-cameras, augmented as proposed here for COCO, could be beneficial for dealing with sharp or blurred target objects that are partially outside a typical camera's field of view. The impact of our approach could be especially helpful in particular applications, such as drone-based following, where even brief interruptions in tracking can ruin a film.

\textbf{Acknowledgements} The authors would like to thank Matthew Johnson and Peter Hedman for their helpful and informative suggestions and discussions. Mohamed is funded by a Microsoft Research PhD Scholarship. 

{\small
\bibliographystyle{ieee_fullname}
\bibliography{egbib}
}

\clearpage
\pagebreak
\onecolumn

\null
\vskip .375in
\begin{center}
  {\Large \bf Improved Handling of Motion Blur in Online Object Detection\\

Supplementary Material \par}
  \vspace*{24pt}

  \vskip .5em
  \vspace*{12pt}
\end{center}

\setcounter{equation}{0}
\setcounter{section}{0}
\setcounter{figure}{0}
\setcounter{table}{0}
\setcounter{page}{1}


\section{Blur Discretization and Blur Space Segmentation}

\subsection{Discretization}

In the main paper, $P$ and $E$ are held to discrete values. 
$P_{1-3}$ are $[0.005, 0.001, 0.00005]$, where a lower value for $P$, $P_3$ for example, gives a more rectilinear blur. Since there are underlying random factors initialized for every blur kernel that are only influenced by $P$, some overlap exists between the type of blur kernels produced across different $P$s. Exposures $E_{1-5}$ are $[1/25, 1/10, 1/5, 1/2, 1]$. Note that for blur trained networks, we don't resize images to a canonical size before blurring; this acts as a mild regularizer and helps creates specialists that are flexible across a range of blur levels.

All mAP@50 scores are reported on the COCO minival set (5000 images). We use a fixed seed for every evaluation when generating blur kernels. 

\begin{figure}[h]
\begin{center}
   \includegraphics[width=1\linewidth]{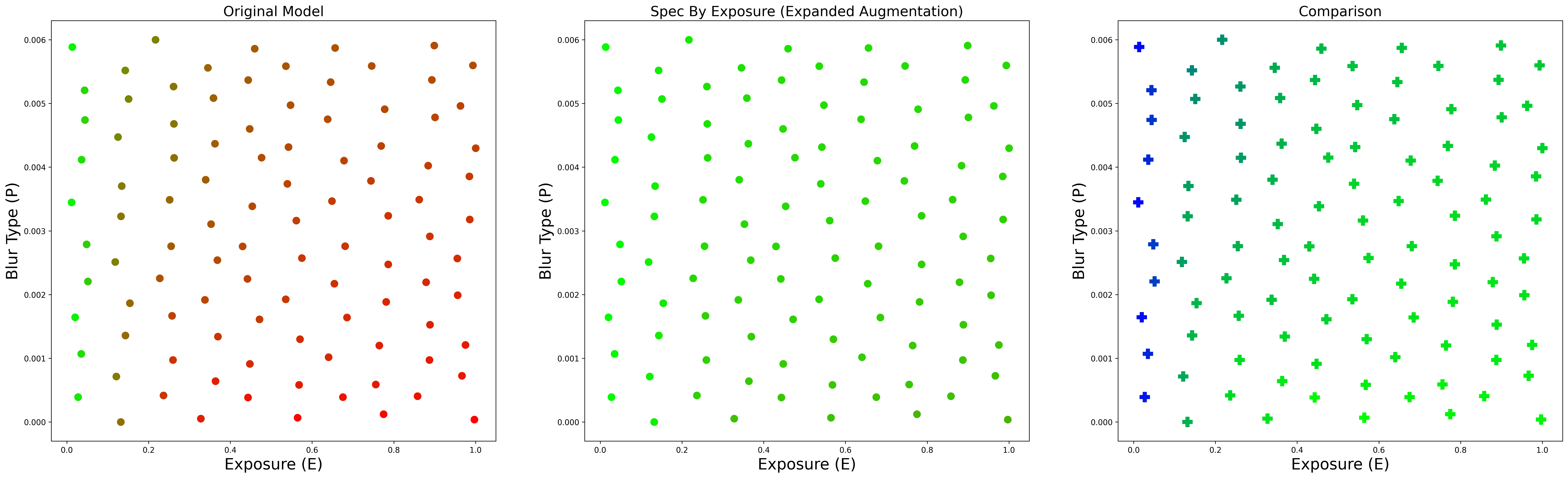}
\end{center}
\caption{Comparison of the original model (ResNet50FPN trained on COCO) and our best model evaluated on expanded labels across a random selection of $P$ and $E$ values. Each marker is a representation of the accuracy (mAP@50) on an evaluation of 2000 images from the COCO minival. For the first two graphs (left to right), the greener the marker the closer it is to an mAP@50 of 61\%. The redder it is, the closer it is to an mAP@50 of 0\%. For the third graph, we visualize the difference between both networks; the greener the marker the larger the diffrence in mAP@50 between our best solution and the original network. The bluer the marker the less the difference is in accuracy. Naturally, at lower exposures, the original network holds up well, but as the exposure is ramped up, and particularly with more rectilinear blur (low $P$ value), the difference is much larger.}
\label{fig:paramSweep}
\end{figure}

While our proposed model was trained with those discrete blur settings, the space of camera-induced blur is not so neatly quantized. To explore a larger cross-section of the continuous blur space, we evaluated a sweep across a random selection of exposures (horizontal axis) and blur types (vertical axis), comparing the original network against our \textbf{Specialized by Exposure Expanded Labels}. Each marker plotted in Fig.~\ref{fig:paramSweep} is an evaluation on 2000 images from the COCO minival set. It visually summarizes that for sharp and barely-blurred images, our approach is negligibly better than the original model. But for essentially all other settings of induced motion blur, our model does measurably better.

\subsection{Segmenting Blur Space: By Type \vs By Exposure}
\begin{figure*}[t]
\begin{center}
\includegraphics[width=1.0\linewidth]{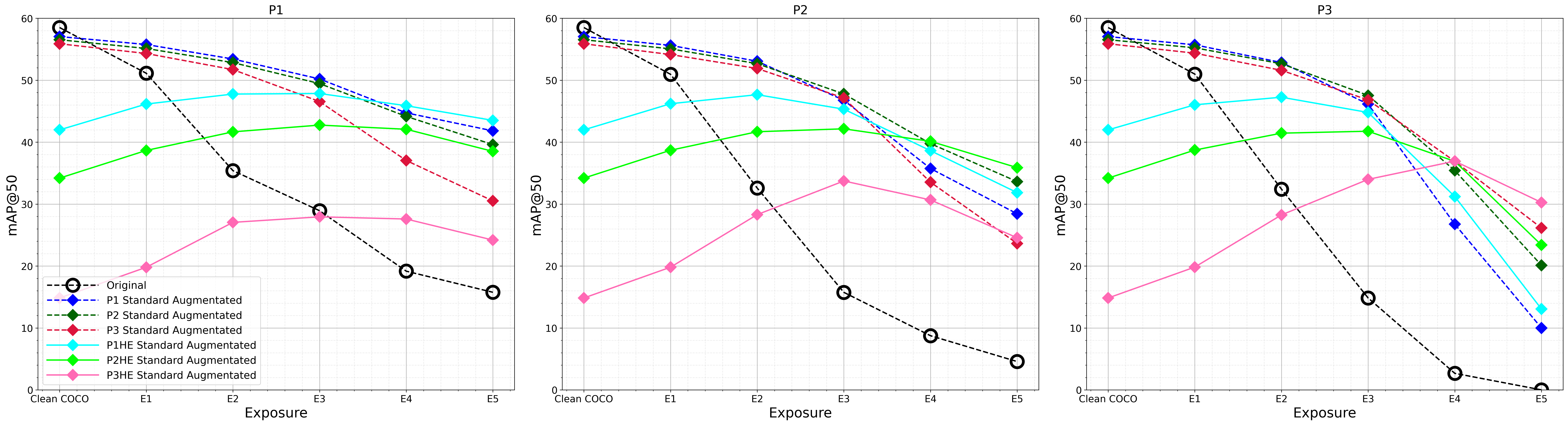}
\end{center}
   \caption{Standard augmented specialized networks performance across different blur types and exposures, evaluated on standard labels.}
\label{fig:specializedPerf}
\end{figure*}

\begin{figure*}[t]
\begin{center}
\includegraphics[width=1.0\linewidth]{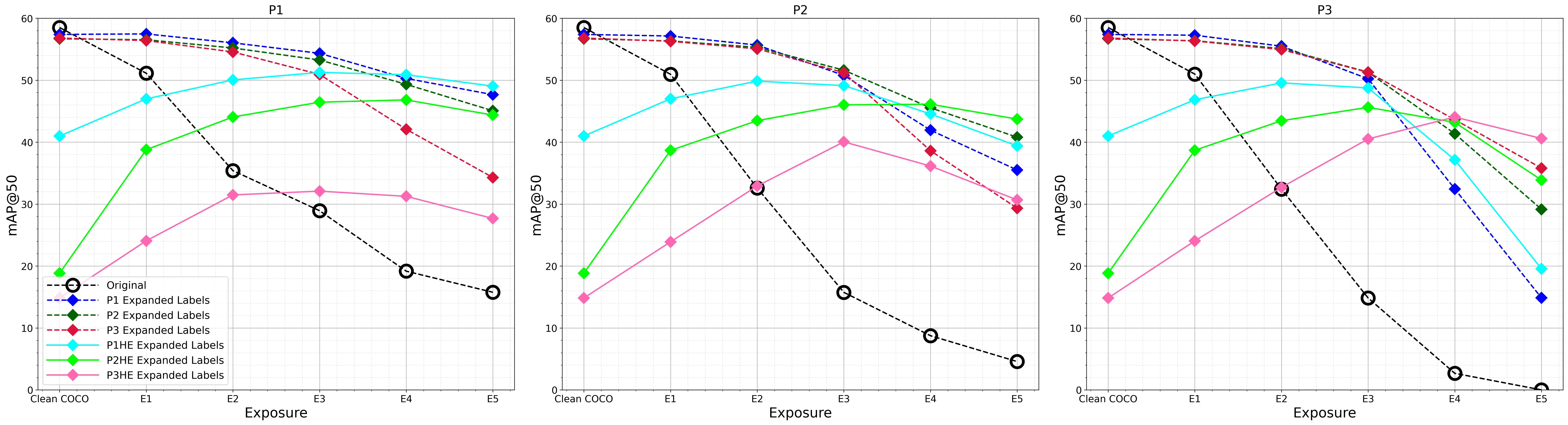}
\end{center}
   \caption{Expanded augmented specialized networks performance across different blur types and exposures, evaluated on expanded labels. Note how here and in Fig.~\ref{fig:specializedPerf}, the each blur type specialized network tends to be better than its peers, especially at higher exposures. The high exposure HE networks outperform the rest at their respective blur type specialty at high exposures.}
\label{fig:specializedPerfExpanded}
\end{figure*}

All general augmented networks (non-specialized) are trained with a mixture of sharp COCO images (10\%) and a random selection of blurry images across $P_{1-3}$ and $E_{1-5}$ (90\%). \textbf{Spec by Type} networks are also trained on the same ratio, but are fixed to a specific $P$. The low exposure network in \textbf{Spec by Exposure} is trained on 25\% sharp images and 75\% blurry images from $P_{1-3}$ and $E_{1-3}$; the three others are trained on 100\% blurry images exclusively from a a specific $P$ and $E_{4,5}$. The performance of these networks separately across blur levels is displayed in Fig.~\ref{fig:specializedPerf} and Fig.~\ref{fig:specializedPerfExpanded}.

\subsection{Blur Estimators}
We use two flavors of a ResNet18 classification network, one for each type of bag of specialists. For Spec By Type, the blur estimator is trained to classify an image as belonging to one of 16 classes, clean or one of $\{P_{1-3} \times E_{1-5}\}$; it achieves 84\% accuracy. For Spec by Exp, images are classified to one of four classes - clean and exposures in $\{E_{1-3}\}$, $\{P_1 \times E_{4,5}\}$, $\{P_2 \times E_{4,5}\}$, or $\{P_3 \times E_{4,5}\}$; this flavor achieves 93\% accuracy. 

Blur estimators are trained for 12 epochs with an initial learning rate of 0.02 (20 images) attenuated by a factor of 10 for each epoch in [3, 7, 10]. Unlike blur augmentation for detection, we resize images to a canonical size of $1333 \times 800$ before blurring.

\section{Zero Centering Ablation}

\begin{figure}[h]
\begin{center}
   \includegraphics[width=0.5\linewidth]{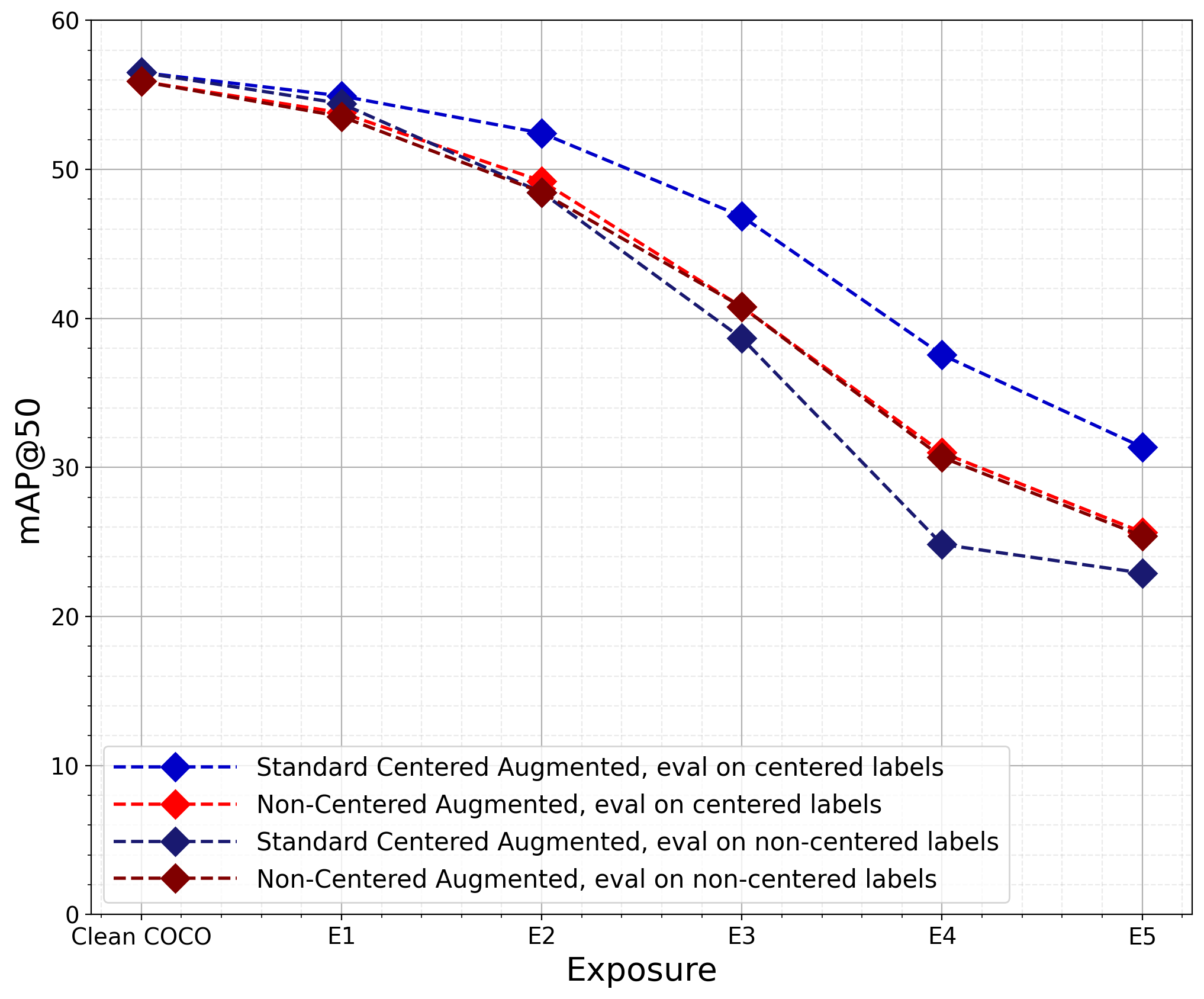}
\end{center}
\caption{ Comparison of different training and evaluation strategies. Results are averaged across the blur types $P_{1-3}$. Evaluating and training on kernels aligned to detection labels (Standard Augmentation and centered labels) scores best.}
\label{fig:t0CompareAll}
\end{figure}

We show how kernel/label centering improves training and test-time accuracy. The main paper features results of evaluating on centered labels that match the barycenter of the kernel. In Fig.~\ref{fig:t0CompareAll} we evaluate on non-aligned kernels and labels as well. Training and evaluating on centered kernels aligned to detection labels produces better scores, likely because the typically non-centered kernels are offset relative to the training bounding boxes. The Non-Centered model achieves the same scores when evaluated with and without centered kernels, indicating that the network has likely learned to find a vague localization and misses boxes altogether that it ought to have detected.

\section{Expanded Labels and What the Network Outputs}
\begin{figure}[h]
\begin{center}
   \includegraphics[width=0.9\linewidth]{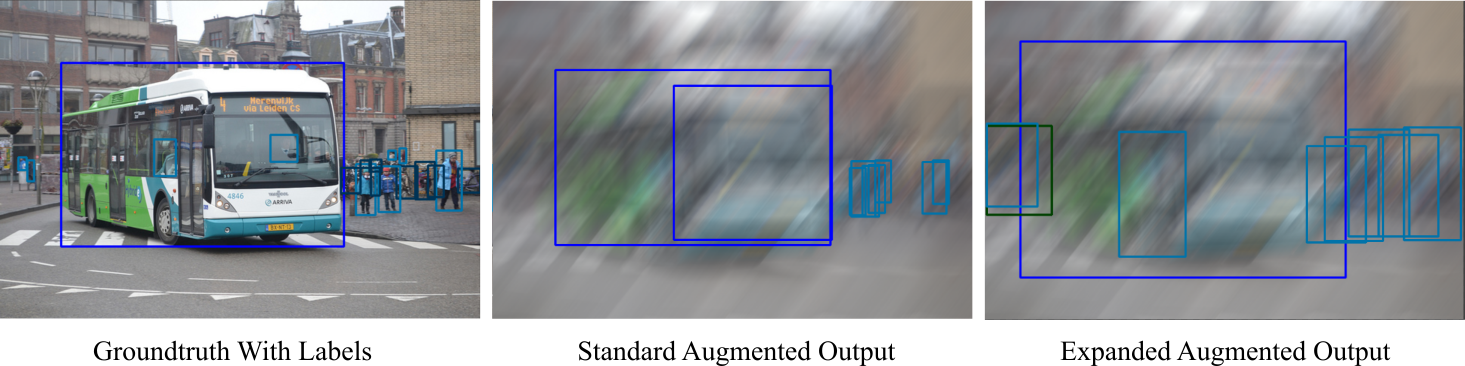}
\end{center}
\caption{Left: Groundtruth image with COCO labels. Middle: Network output from the Standard Augment network. Right: Network output from the Expanded Augment network. Expanded augment networks learn to output boxes that represent the superset of all locations an object has been at during an exposure.}
\label{fig:expandedoutputexample}
\end{figure}

Fig.~\ref{fig:expandedoutputexample} shows an example image with motion blur, and the output from both the \textbf{Standard Augmented} and the \textbf{Expanded Augmented} networks. The Expanded augmented network learns to predict bounding boxes that capture the superset of all spatial locations an object occupied during an exposure. This seems to be an easier objective for the network to learn. While one could argue that downstream tasks may prefer original-sized bounding boxes as shown computed by the Standard Augmented model, there is no good compromise there: the middle of the blurred object could be a ``stale'' image-space location compared to where the object is at the end of the exposure, in, \eg a tracking-by-detection task. In qualitative examples, the expanded networks manage to detect bounding boxes that are otherwise missed by their standard counterparts, see Fig.\ref{fig:expandedoutputexample}.

\section{Minibatch Normalization as Schneider~\etal~\cite{schneider2020improvingCovariantShift}}

\begin{figure}[h]
\begin{center}
   \includegraphics[width=0.5\linewidth]{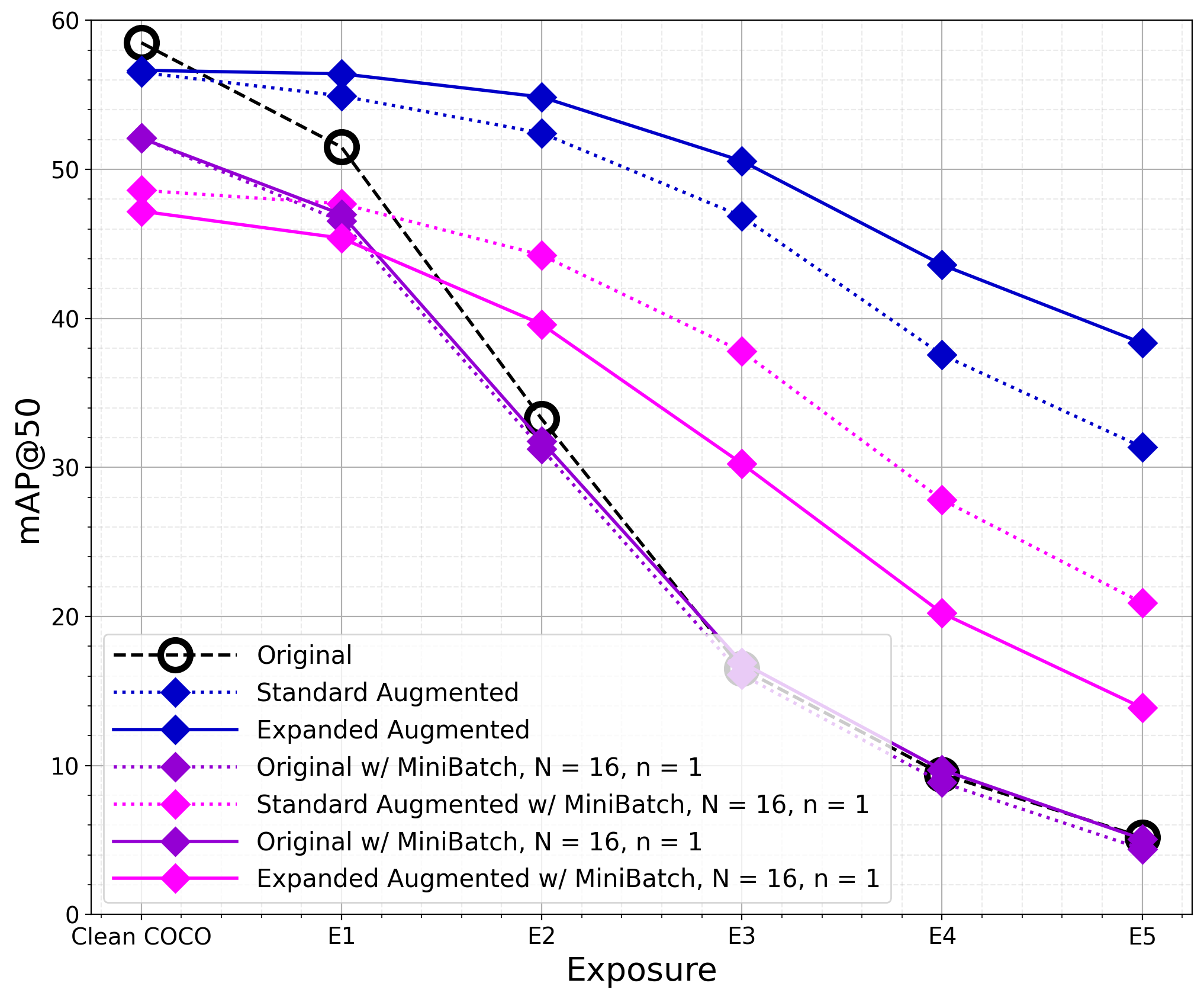}
\end{center}
\caption{Comparison of using minibatch normalization on both the original network and blur augmented networks. For only this graph: solid lines are evaluation runs on expanded labels and dashed lines are evaluated on standard labels; the exception here is the original model which is evaluated on expanded labels. Results are averaged across $P_{1-3}$. }
\label{fig:minibatchComplete}
\end{figure}

In this late-breaking NeurIPS 2020 paper, results are reported for minibatch normalization on networks already trained with augmentation for blurry images. As per their algorithm, we perform minibatch normalization by finding the statistics of the activations of an input example, $\mu$ and $\sigma$, and computing a weighted sum with the training statistics using $N=16$ and $n=1$. This is done progressively in one forward pass. In Fig.~\ref{fig:minibatchComplete}, results are reported for the performance of the original model with this modification. We also experimented with finding an accurate estimate of the target distribution for blurry images by running a large portion of the train set under a specific type of blur and exposure as many times as there are batchnorm layers in the network with $n = 2048$, and using that as normalization statistics, but this did not constitute an improvement. 
Despite the appeal of this test-time approach, object detection was not substantially better off with it, so we excluded it from our final model.

\section{Defocus And Motion Blur}
\begin{figure}[h]
\begin{center}
\includegraphics[width=0.5\linewidth]{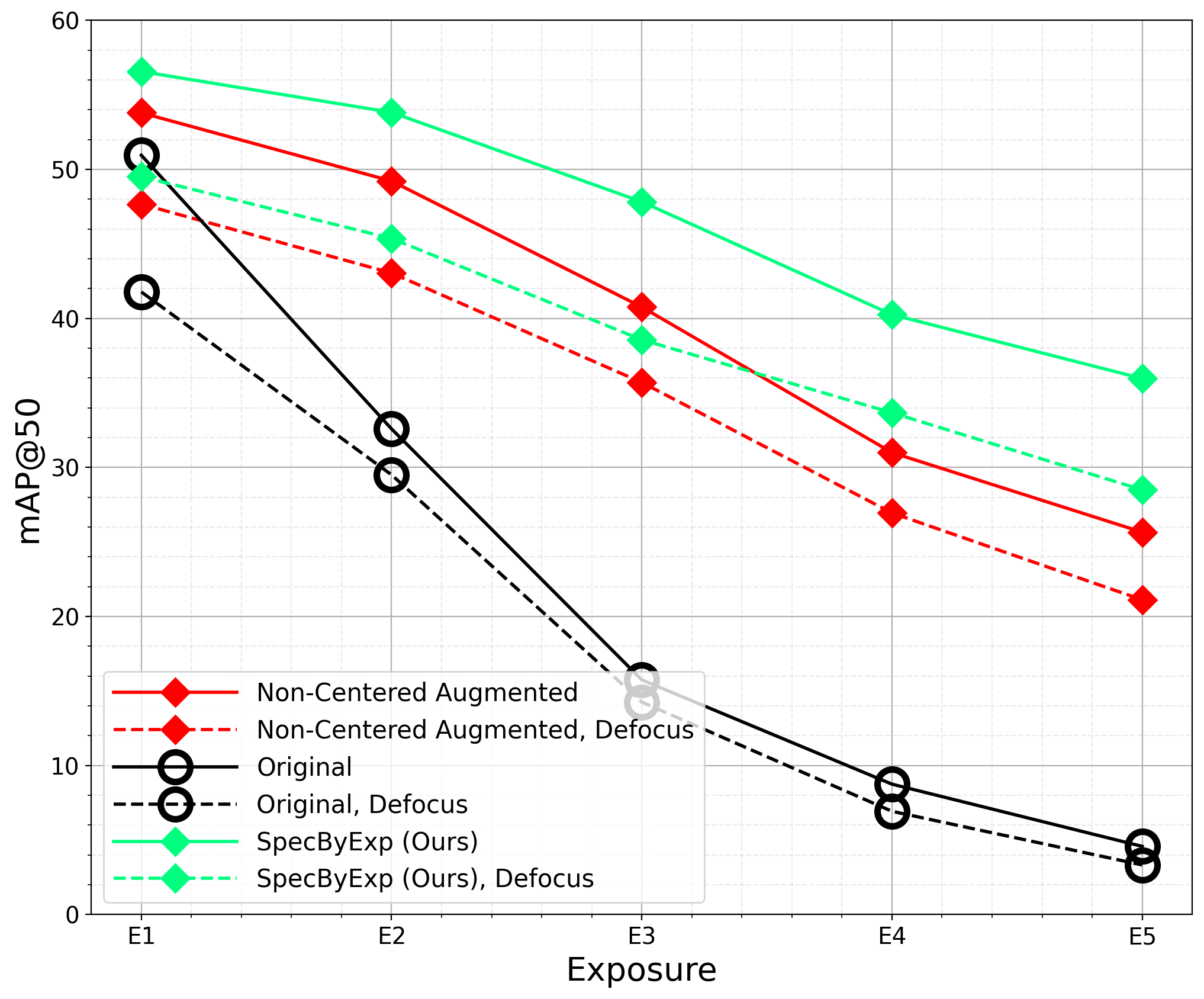}
\end{center}
\caption{Models evaluated on motion blur with and without defocusing. $E$ is motion blur extent; each point averages blur types $P_{1-3}$. Defocus is simulated by Gaussian blurring each motion-blur kernel with a random $\sigma$. Defocus hurts, but our model still performs well, especially compared to no-centering and the original network.}
\label{fig:defocusComp} 
\end{figure}

Our models are more resilient to camera defocus blur than the original. $P_{1}$ is close to simulating defocus blur since the camera trajectory loops in place. We Gaussian blur each motion-blur kernel with a random $\sigma$ across all blur types and exposures. We report results in Fig.~\ref{fig:defocusComp}.

\section{Real-World Blur Datasets}
We evaluate our models on on two pseudo-real blur datasets, GOPRO~\cite{nah2017deep} and REDS~\cite{Nah_2019_CVPR_Workshops_REDS}, and a real-world blur dataset, RealBlur~\cite{rim_2020_ECCV}, obtained using shutter tied cameras. These datasets don't have box annotations, so we utilize a state-of-the-art high accuracy detector, DetectoRS~\cite{detectors}, to obtain pseudo-groundtruth bounding-boxes for evaluation. For evaluating expanded bounding boxes, we generate our own GOPRO testset using grountruth sharp frames and use flow computed using ~\cite{teed2020raft} for bounding-box expansion.

We stick to the canonical train and test sets when available. However when either the train and sets combined don't contain enough images for reliable evaluation or when we need to estimate flow on source high-frame rate images, we synthesize our own set of blurry images using sharp images from the datasets. 

We perform standard evaluation on RealBlur, GOPRO, and REDS. RealBlur sharp and blurry frames are naturally aligned during capture and the alignment is further refined in the post process described in the paper~\cite{rim_2020_ECCV}. GOPRO and REDS report a sharp frame as one in the middle of the window to synthesize a blurry frame. Although this doesn't necessarily equate to centering the blur kernel since movement can be asymmetric on either side of the sharp frame, we use this as our ``standard'' evaluation as it's the closest approximation given the data.

For RealBlur, we use both train and test sets (4,738 pairs) and set to the confidence threshold on pseudo-groundtruth boxes to 0.6. For REDS, we sample 5,000 frames from the train and validation, set the confidence threshold to 0.4, and allow only a maximum of 20 images without bounding boxes. For GOPRO, we use the combination of the train and test sets (3214 pairs) and set the confidence threshold to 0.4. 

For expanded evaluation, we synthesize 5,442 blurry GOPRO frames using the method and window size limits outlined in~\cite{nah2017deep}, and we set the confidence threshold on pseudo-groundtruth bounding boxes to 0.6. We omit empty scenes with no COCO object classes, namely \verb|GOPR0374_11_00|, \verb|GOPR0374_11_01|, \verb|GOPR0374_11_02|, \verb|GOPR0374_11_03|, \verb|GOPR0857_11_00|, \verb|GOPR0868_11_00|, \verb|GOPR0868_11_02|, \verb|GOPR0871_11_00|, and \verb|GOPR0396_11_00|. During this process, we note the sharp frames used to synthesize blurry frames and obtain low-resolution flow fields using RAFT~\cite{teed2020raft} to estimate where objects have moved during the exposure. We use low-resolution as apposed to the refined maps to avoid artefacts at object boundaries. We advect each bounding box corner using the estimated flow fields both forwards and backwards on either side of the sharp frame, stopping at the assigned blur window size. We then assign the bounding box corners to the super-set of both the original points and the advected points. These new boxes are estimates of the super-set location of where an object has been in an exposure, and are used during expanded evaluation.

\section{Qualitative Results}
You can find a video with qualitative results and a visual explanation of our method \url{visual.cs.ucl.ac.uk/pubs/handlingMotionBlur/}. 

There, we show real world examples where our model, based on the two proposed remedies, manages to detect objects in many places where the original model fails, especially when the ratio of camera motion to object size is high. Following on from the quantitative experiments in the paper and here, we synthesize blurry COCO images (in the same spirit as~\cite{michaelis2019benchmarking, hendrycks2019benchmarking}) and show sample results in the video.

\section{Results Tables}
Table~\ref{table:nonexpand} and Table~\ref{table:expand} contain the raw results used to generate Fig.~\ref{fig:nonExpandAccuracyValues} and Fig.~\ref{fig:expandAccuracyValues} in the paper. Table~\ref{table:nonexpandSpecialists} and Table~\ref{table:expandSpecialists} show raw numbers for generating Fig.~\ref{fig:specializedPerf} and Fig.~\ref{fig:specializedPerfExpanded}. 

\begin{table}
\begin{center}

\begin{tabular}{|l||c|c|c|c|c|c|}
\hline
Variant & Clean & $E_1$ & $E_2$ & $E_3$ & $E_4$ & $E_5$ \\
\hline
Original & 58.50 & 50.95 & 32.59 & 15.75 & 8.75 & 4.58 \\ 
\hline
Deblur then Original & 55.50 & 49.18 & 42.13 & 30.31 & 12.72 & 6.26 \\ 
Deblur then Standard Augmented & 53.90 & 51.47 & 48.44 & 40.35 & 23.85 & 15.86 \\ 
Squint & 55.65 & 54.30 & 51.76 & 46.21 & 37.24 & 31.39 \\ 
AugMix (Non Expanded) & \textbf{59.34} & 53.13 & 38.07 & 20.70 & 13.63 & 8.21 \\ 
AugMix PixelLevel & \underline{58.93} & 51.68 & 32.10 & 14.84 & 9.12 & 4.48 \\ 
Original w/ MiniBatch, N = 16, n = 1 & 52.10 & 46.53 & 31.25 & 16.10 & 8.86 & 4.40 \\ 
Standard Augmented w/ MiniBatch, N = 16, n = 1 & 48.60 & 47.70 & 44.25 & 37.79 & 27.84 & 20.92 \\ 
Non-Centered Augmented & 55.91 & 53.80 & 49.22 & 40.77 & 31.00 & 25.66 \\ 
Standard Augmented w/ NonSpatial Augmix & 55.77 & 54.15 & 51.95 & 46.53 & 38.41 & 31.67 \\ 
Standard Augmented & 56.51 & 54.93 & 52.44 & 46.85 & 37.56 & 31.37 \\ 
Spec By Type & 56.50 & \underline{55.39} & \underline{52.33} & \textbf{47.78} & \textbf{39.81} & \underline{33.84} \\ 
Spec By Exposure (Ours) & 58.55 & \textbf{56.57} & \textbf{53.83} & \underline{47.74} & \underline{40.21} & \textbf{35.93} \\ 
\hline
\end{tabular}

\begin{tabular}{|l||c|c|c|c|c|c|}
\hline
Variant & Clean & $E_1$ & $E_2$ & $E_3$ & $E_4$ & $E_5$ \\
\hline
Original & 58.50 & 50.95 & 32.59 & 15.75 & 8.75 & 4.58 \\ 
\hline
Deblur then Original & 55.50 & 49.18 & 42.13 & 30.31 & 12.72 & 6.26 \\ 
Deblur then Standard Augmented & 53.90 & 51.47 & 48.44 & 40.35 & 23.85 & 15.86 \\ 
Squint & 55.65 & 54.30 & 51.76 & 46.21 & 37.24 & 31.39 \\ 
AugMix (Non Expanded) & \textbf{59.34} & 53.13 & 38.07 & 20.70 & 13.63 & 8.21 \\ 
AugMix PixelLevel & \underline{58.93} & 51.68 & 32.10 & 14.84 & 9.12 & 4.48 \\ 
Original w/ MiniBatch, N = 16, n = 1 & 52.10 & 46.53 & 31.25 & 16.10 & 8.86 & 4.40 \\ 
Standard Augmented w/ MiniBatch, N = 16, n = 1 & 48.60 & 47.70 & 44.25 & 37.79 & 27.84 & 20.92 \\ 
Non-Centered Augmented & 55.91 & 53.80 & 49.22 & 40.77 & 31.00 & 25.66 \\ 
Standard Augmented w/ NonSpatial Augmix & 55.77 & 54.15 & 51.95 & 46.53 & 38.41 & 31.67 \\ 
Standard Augmented & 56.51 & 54.93 & \underline{52.44} & 46.85 & 37.56 & 31.37 \\ 
Spec By Type & 56.50 & \underline{55.39} & 52.33 & \textbf{47.78} & \underline{39.81} & \underline{33.84} \\ 
Spec By Exposure (Ours) & 58.55 & \textbf{56.57} & \textbf{53.83} & \underline{47.74} & \textbf{40.21} & \textbf{35.93} \\ 
\hline
\end{tabular}

\begin{tabular}{|l||c|c|c|c|c|c|}
\hline
Variant & Clean & $E_1$ & $E_2$ & $E_3$ & $E_4$ & $E_5$ \\
\hline
Original & 58.50 & 50.95 & 32.59 & 15.75 & 8.75 & 4.58 \\ 
\hline
Deblur then Original & 55.50 & 49.18 & 42.13 & 30.31 & 12.72 & 6.26 \\ 
Deblur then Standard Augmented & 53.90 & 51.47 & 48.44 & 40.35 & 23.85 & 15.86 \\ 
Squint & 55.65 & 54.30 & 51.76 & 46.21 & 37.24 & 31.39 \\ 
AugMix (Non Expanded) & \textbf{59.34} & 53.13 & 38.07 & 20.70 & 13.63 & 8.21 \\ 
AugMix PixelLevel & \underline{58.93} & 51.68 & 32.10 & 14.84 & 9.12 & 4.48 \\ 
Original w/ MiniBatch, N = 16, n = 1 & 52.10 & 46.53 & 31.25 & 16.10 & 8.86 & 4.40 \\ 
Standard Augmented w/ MiniBatch, N = 16, n = 1 & 48.60 & 47.70 & 44.25 & 37.79 & 27.84 & 20.92 \\ 
Non-Centered Augmented & 55.91 & 53.80 & 49.22 & 40.77 & 31.00 & 25.66 \\ 
Standard Augmented w/ NonSpatial Augmix & 55.77 & 54.15 & 51.95 & 46.53 & 38.41 & 31.67 \\ 
Standard Augmented & 56.51 & 54.93 & \underline{52.44} & 46.85 & 37.56 & 31.37 \\ 
Spec By Type & 56.50 & \underline{55.39} & 52.33 & \underline{47.78} & \underline{39.81} & \underline{33.84} \\ 
Spec By Exposure (Ours) & 58.55 & \textbf{56.57} & \textbf{53.83} & \textbf{47.74} & \textbf{40.21} & \textbf{35.93} \\ 
\hline
\end{tabular}

\end{center}
\caption{Raw numbers from Fig.~\ref{fig:nonExpandAccuracyValues} in the paper. Non-expanded labels used during evaluation. Results are on the COCO minival set under different blur parameters and exposure. From top to bottom, the blur type changes from $P_1$ to $P_2$ to $P_3$. Networks trained with blur augmentation would be trained on non-expanded labels.}
\label{table:nonexpand}
\end{table}

\begin{table}
\begin{center}

\begin{tabular}{|l||c|c|c|c|c|c|}
\hline
Variant & Clean & $E_1$ & $E_2$ & $E_3$ & $E_4$ & $E_5$ \\
\hline
Original & 58.50 & 51.50 & 33.26 & 16.49 & 9.41 & 5.20 \\ 
\hline
Deblur then Original & 55.50 & 49.50 & 41.07 & 28.32 & 12.19 & 6.24 \\ 
Squint Expanded Labels & 56.15 & 56.25 & 54.53 & 50.09 & 42.92 & 37.66 \\ 
AugMix Expanded Labels & 51.80 & 46.62 & 34.21 & 18.99 & 11.32 & 6.15 \\ 
AugMix PixelLevel & \textbf{58.93} & 51.50 & 33.26 & 16.49 & 9.41 & 0.05 \\ 
Original w/ MiniBatch, N = 16, n = 1 & 52.10 & 46.99 & 31.77 & 16.92 & 9.71 & 5.07 \\ 
Expanded Labels w/ MiniBatch, N = 16, n = 1 & 47.20 & 45.39 & 39.61 & 30.26 & 20.23 & 13.88 \\ 
Expanded Labels & 56.65 & 56.42 & 54.86 & 50.57 & 43.60 & 38.35 \\ 
Expanded Labels w/ NonSpatial Augmix & 56.33 & 55.99 & 54.54 & 50.32 & 43.10 & 37.85 \\ 
Spec By Type Expanded Labels & 56.70 & \underline{56.75} & \underline{55.23} & \textbf{51.59} & \underline{45.55} & \underline{40.81} \\ 
Spec By Exposure Expanded Labels (Our Best) & \underline{58.62} & \textbf{58.01} & \textbf{56.40} & \underline{50.97} & \textbf{46.37} & \textbf{43.78} \\ 
\hline
\end{tabular}

\begin{tabular}{|l||c|c|c|c|c|c|}
\hline
Variant & Clean & $E_1$ & $E_2$ & $E_3$ & $E_4$ & $E_5$ \\
\hline
Original & 58.50 & 51.50 & 33.26 & 16.49 & 9.41 & 5.20 \\ 
\hline
Deblur then Original & 55.50 & 49.50 & 41.07 & 28.32 & 12.19 & 6.24 \\ 
Squint Expanded Labels & 56.15 & 56.25 & 54.53 & 50.09 & 42.92 & 37.66 \\ 
AugMix Expanded Labels & 51.80 & 46.62 & 34.21 & 18.99 & 11.32 & 6.15 \\ 
AugMix PixelLevel & \textbf{58.93} & 51.50 & 33.26 & 16.49 & 9.41 & 0.05 \\ 
Original w/ MiniBatch, N = 16, n = 1 & 52.10 & 46.99 & 31.77 & 16.92 & 9.71 & 5.07 \\ 
Expanded Labels w/ MiniBatch, N = 16, n = 1 & 47.20 & 45.39 & 39.61 & 30.26 & 20.23 & 13.88 \\ 
Expanded Labels & 56.65 & 56.42 & 54.86 & 50.57 & 43.60 & 38.35 \\ 
Expanded Labels w/ NonSpatial Augmix & 56.33 & 55.99 & 54.54 & 50.32 & 43.10 & 37.85 \\ 
Spec By Type Expanded Labels & 56.70 & \underline{56.75} & \underline{55.23} & \textbf{51.59} & \underline{45.55} & \underline{40.81} \\ 
Spec By Exposure Expanded Labels (Our Best) & \underline{58.62} & \textbf{58.01} & \textbf{56.40} & \underline{50.97} & \textbf{46.37} & \textbf{43.78} \\ 
\hline
\end{tabular}

\begin{tabular}{|l||c|c|c|c|c|c|}
\hline
Variant & Clean & $E_1$ & $E_2$ & $E_3$ & $E_4$ & $E_5$ \\
\hline
Original & 58.50 & 51.50 & 33.26 & 16.49 & 9.41 & 5.20 \\ 
\hline
Deblur then Original & 55.50 & 49.50 & 41.07 & 28.32 & 12.19 & 6.24 \\ 
Squint Expanded Labels & 56.15 & 56.25 & 54.53 & 50.09 & 42.92 & 37.66 \\ 
AugMix Expanded Labels & 51.80 & 46.62 & 34.21 & 18.99 & 11.32 & 6.15 \\ 
AugMix PixelLevel & \textbf{58.93} & 51.50 & 33.26 & 16.49 & 9.41 & 0.05 \\ 
Original w/ MiniBatch, N = 16, n = 1 & 52.10 & 46.99 & 31.77 & 16.92 & 9.71 & 5.07 \\ 
Expanded Labels w/ MiniBatch, N = 16, n = 1 & 47.20 & 45.39 & 39.61 & 30.26 & 20.23 & 13.88 \\ 
Expanded Labels & 56.65 & 56.42 & 54.86 & 50.57 & 43.60 & 38.35 \\ 
Expanded Labels w/ NonSpatial Augmix & 56.33 & 55.99 & 54.54 & 50.32 & 43.10 & 37.85 \\ 
Spec By Type Expanded Labels & 56.70 & \underline{56.75} & \underline{55.23} & \textbf{51.59} & \underline{45.55} & \underline{40.81} \\ 
Spec By Exposure Expanded Labels (Our Best) & \underline{58.62} & \textbf{58.01} & \textbf{56.40} & \underline{50.97} & \textbf{46.37} & \textbf{43.78} \\ 
\hline
\end{tabular}

\end{center}
\caption{Raw numbers from Fig.~\ref{fig:expandAccuracyValues} in the paper. Expanded labels used during evaluation. Results are on the COCO minival set under different blur parameters and exposure. From top to bottom, the blur type changes from $P_1$ to $P_2$ to $P_3$. 
}
\label{table:expand}
\end{table}

\begin{table}
\begin{center}

\begin{tabular}{|l||c|c|c|c|c|c|}
\hline
Variant & Clean & $E_1$ & $E_2$ & $E_3$ & $E_4$ & $E_5$ \\
\hline
Original & \underline{58.50} & 50.95 & 32.59 & 15.75 & 8.75 & 4.58 \\ 
Low-Exposure Augmented & \textbf{58.55} & \textbf{56.57} & \textbf{53.83} & \textbf{47.83} & 29.32 & 18.64 \\ 
P1 Standard Augmentated & 57.04 & \underline{55.62} & \underline{53.06} & \underline{46.76} & \underline{35.74} & \underline{28.43} \\ 
P2 Standard Augmentated & 56.53 & 55.06 & 52.67 & 47.78 & 39.76 & 33.63 \\ 
P3 Standard Augmentated & 55.88 & 54.14 & 51.89 & 47.12 & 33.53 & 23.64 \\ 
P1HE Standard Augmentated & 41.98 & 46.20 & 47.66 & 45.32 & \textbf{38.62} & \textbf{31.85} \\ 
P2HE Standard Augmentated & 34.19 & 38.69 & 41.68 & 42.15 & 40.13 & 35.88 \\ 
P3HE Standard Augmentated & 14.84 & 19.81 & 28.30 & 33.73 & 30.67 & 24.56 \\ 
\hline
\end{tabular}

\begin{tabular}{|l||c|c|c|c|c|c|}
\hline
Variant & Clean & $E_1$ & $E_2$ & $E_3$ & $E_4$ & $E_5$ \\
\hline
Original & \underline{58.50} & 50.95 & 32.59 & 15.75 & 8.75 & 4.58 \\ 
Low-Exposure Augmented & \textbf{58.55} & \textbf{56.57} & \textbf{53.83} & \textbf{47.83} & 29.32 & 18.64 \\ 
P1 Standard Augmentated & 57.04 & \underline{55.62} & \underline{53.06} & 46.76 & 35.74 & 28.43 \\ 
P2 Standard Augmentated & 56.53 & 55.06 & 52.67 & \underline{47.78} & \underline{39.76} & \underline{33.63} \\ 
P3 Standard Augmentated & 55.88 & 54.14 & 51.89 & 47.12 & 33.53 & 23.64 \\ 
P1HE Standard Augmentated & 41.98 & 46.20 & 47.66 & 45.32 & 38.62 & 31.85 \\ 
P2HE Standard Augmentated & 34.19 & 38.69 & 41.68 & 42.15 & \textbf{40.13} & \textbf{35.88} \\ 
P3HE Standard Augmentated & 14.84 & 19.81 & 28.30 & 33.73 & 30.67 & 24.56 \\ 
\hline
\end{tabular}

\begin{tabular}{|l||c|c|c|c|c|c|}
\hline
Variant & Clean & $E_1$ & $E_2$ & $E_3$ & $E_4$ & $E_5$ \\
\hline
Original & \underline{58.50} & 50.95 & 32.59 & 15.75 & 8.75 & 4.58 \\ 
Low-Exposure Augmented & \textbf{58.55} & \textbf{56.57} & \textbf{53.83} & \textbf{47.83} & 29.32 & 18.64 \\ 
P1 Standard Augmentated & 57.04 & \underline{55.62} & \underline{53.06} & 46.76 & 35.74 & 28.43 \\ 
P2 Standard Augmentated & 56.53 & 55.06 & 52.67 & \underline{47.78} & 39.76 & 33.63 \\ 
P3 Standard Augmentated & 55.88 & 54.14 & 51.89 & 47.12 & \underline{33.53} & \underline{23.64} \\ 
P1HE Standard Augmentated & 41.98 & 46.20 & 47.66 & 45.32 & 38.62 & 31.85 \\ 
P2HE Standard Augmentated & 34.19 & 38.69 & 41.68 & 42.15 & 40.13 & 35.88 \\ 
P3HE Standard Augmentated & 14.84 & 19.81 & 28.30 & 33.73 & \textbf{30.67} & \textbf{24.56} \\ 
\hline
\end{tabular}

\end{center}
\caption{Raw numbers for standard augmented specialists performance (Fig.~\ref{fig:specializedPerf}). Results are on the COCO minival set under different blur parameters and exposure. From top to bottom, the blur type changes from $P_1$ to $P_2$ to $P_3$. Networks are trained and evaluated on non-expanded ``standard'' labels under blur augmentation, with the exception of Original.}

\label{table:nonexpandSpecialists}
\end{table}

\begin{table}
\begin{center}

\begin{tabular}{|l||c|c|c|c|c|c|}
\hline
Variant & Clean & $E_1$ & $E_2$ & $E_3$ & $E_4$ & $E_5$ \\
\hline
Original & \underline{58.50} & 51.50 & 33.26 & 16.49 & 9.41 & 5.20 \\
Low-Exposure Expanded Labels & \textbf{58.62} & \textbf{58.06} & \textbf{56.38} & 50.97 & 33.95 & 22.40 \\ 
P1 Expanded Labels & 57.39 & \underline{57.13} & \underline{55.67} & \textbf{50.78} & \underline{41.93} & \underline{35.50} \\ 
P2 Expanded Labels & 56.68 & 56.34 & 55.30 & \underline{51.62} & 45.54 & 40.80 \\ 
P3 Expanded Labels & 56.80 & 56.28 & 55.06 & 51.22 & 38.61 & 29.34 \\ 
P1HE Expanded Labels & 40.99 & 47.00 & 49.85 & 49.13 & \textbf{44.55} & \textbf{39.40} \\ 
P2HE Expanded Labels & 18.84 & 38.67 & 43.47 & 46.02 & 46.12 & 43.72 \\ 
P3HE Expanded Labels & 14.84 & 23.90 & 32.92 & 40.05 & 36.14 & 30.67 \\ 
\hline
\end{tabular}

\begin{tabular}{|l||c|c|c|c|c|c|}
\hline
Variant & Clean & $E_1$ & $E_2$ & $E_3$ & $E_4$ & $E_5$ \\
\hline
Original & \underline{58.50} & 51.50 & 33.26 & 16.49 & 9.41 & 5.20 \\ 
Low-Exposure Expanded Labels & \textbf{58.62} & \textbf{58.06} & \textbf{56.38} & 50.97 & 33.95 & 22.40 \\ 
P1 Expanded Labels & 57.39 & \underline{57.13} & \underline{55.67} & 50.78 & 41.93 & 35.50 \\ 
P2 Expanded Labels & 56.68 & 56.34 & 55.30 & \textbf{51.62} & \underline{45.54} & \underline{40.80} \\ 
P3 Expanded Labels & 56.80 & 56.28 & 55.06 & \underline{51.22} & 38.61 & 29.34 \\ 
P1HE Expanded Labels & 40.99 & 47.00 & 49.85 & 49.13 & 44.55 & 39.40 \\ 
P2HE Expanded Labels & 18.84 & 38.67 & 43.47 & 46.02 & \textbf{46.12} & \textbf{43.72} \\ 
P3HE Expanded Labels & 14.84 & 23.90 & 32.92 & 40.05 & 36.14 & 30.67 \\ 
\hline
\end{tabular}

\begin{tabular}{|l||c|c|c|c|c|c|}
\hline
Variant & Clean & $E_1$ & $E_2$ & $E_3$ & $E_4$ & $E_5$ \\
\hline
Original & \underline{58.50} & 51.50 & 33.26 & 16.49 & 9.41 & 5.20 \\
Low-Exposure Expanded Labels & \textbf{58.62} & \textbf{58.06} & \textbf{56.38} & 50.97 & 33.95 & 22.40 \\ 
P1 Expanded Labels & 57.39 & \underline{57.13} & \underline{55.67} & 50.78 & 41.93 & 35.50 \\ 
P2 Expanded Labels & 56.68 & 56.34 & 55.30 & \textbf{51.62} & 45.54 & 40.80 \\ 
P3 Expanded Labels & 56.80 & 56.28 & 55.06 & \underline{51.22} & \underline{38.61} & \underline{29.34} \\ 
P1HE Expanded Labels & 40.99 & 47.00 & 49.85 & 49.13 & 44.55 & 39.40 \\ 
P2HE Expanded Labels & 18.84 & 38.67 & 43.47 & 46.02 & 46.12 & 43.72 \\ 
P3HE Expanded Labels & 14.84 & 23.90 & 32.92 & 40.05 & \textbf{36.14} & \textbf{30.67} \\ 
\hline
\end{tabular}

\end{center}
\caption{Raw numbers for expanded augmented specialists performance (Fig.~\ref{fig:specializedPerfExpanded}). Results are on the COCO minival set under different blur parameters and exposure. From top to bottom, the blur type changes from $P_1$ to $P_2$ to $P_3$. Networks are trained and evaluated on expanded labels under blur augmentation, with the exception of Original.}

\label{table:expandSpecialists}
\end{table}

\end{document}